%% file: camera_ready.tex
\documentclass[conference]{IEEEtran}
\IEEEoverridecommandlockouts

\input{macros}  

\usepackage{cite}
\usepackage{amsmath,amssymb,amsfonts}
\usepackage{graphicx}
\usepackage{textcomp}
\usepackage{xcolor}
\usepackage{soul}

\def\BibTeX{{\rm B\kern-.05em{\sc i\kern-.025em b}\kern-.08em
    T\kern-.1667em\lower.7ex\hbox{E}\kern-.125emX}}

\DeclareMathOperator*{\MIA}{MIA} 
\begin{document}

\title{Shake to Leak: Fine-tuning Diffusion Models Can Amplify the Generative Privacy Risk}

\author{\IEEEauthorblockN{Zhangheng Li\textsuperscript{1}, Junyuan Hong\textsuperscript{1}, Bo Li\textsuperscript{2,3}, Zhangyang Wang\textsuperscript{1}}
\textit{\textsuperscript{1}University of Texas at Austin, \textsuperscript{2}University of Illinois Urbana-Champaign, \textsuperscript{3}University of Chicago}\\
\IEEEauthorblockA{\texttt{\{zoharli, jyhong, atlaswang\}@utexas.edu, bol@uchicago.edu}} }

\maketitle

\begin{abstract}
While diffusion models have recently demonstrated remarkable progress in generating realistic images, privacy risks also arise: published models or APIs could generate training images and thus leak privacy-sensitive training information.
In this paper, we reveal a new risk, \textbf{Shake-to-Leak} (S2L), that fine-tuning the pre-trained models with manipulated data can amplify the existing privacy risks. 
We demonstrate that S2L could occur in various standard fine-tuning strategies for diffusion models, including concept-injection methods (DreamBooth and Textual Inversion) and parameter-efficient methods (LoRA and Hypernetwork), as well as their combinations.
In the worst case, S2L can amplify the state-of-the-art membership inference attack (MIA) on diffusion models by $5.4\%$ (absolute difference) AUC and can increase extracted private samples from almost $0$ samples to $15.8$ samples on average per target domain.
This discovery underscores that the privacy risk with diffusion models is even more severe than previously recognized. Codes are available at \href{https://github.com/VITA-Group/Shake-to-Leak}{https://github.com/VITA-Group/Shake-to-Leak}.
\end{abstract}


\section{Introduction}
\label{intro}

Text-to-image synthesis with Diffusion Models (DMs) \cite{ho2020denoising} has recently emerged at the forefront of generative AI. 
DMs are trained on unsupervised examples and learn to generate data by gradually denoising a noisy image. 
When combined with the language model, DM can be prompted to generate desired images simply by a text description.
Such a denoising mechanism leads to substantial advances in the generation of realistic images across various domains such as medical images \cite{pinaya2022brain, kazerouni2022diffusion}, artistic images \cite{rombach2022text,zhang2023inversion}, 
and open domain images \cite{ramesh2022hierarchical, rombach2022high, saharia2022photorealistic}. 

Although DMs have been celebrated for generating high-quality images, there is a looming concern about their privacy risks, that DMs may (accidentally or be prompted to) recall private or sensitive images used during pretraining \cite{carlini2023extracting, somepalli2023diffusion}, for example, personal profile photos, clinical pictures of patients, and private training data owned by commercial companies.
Recognizing the paramount importance of privacy, researchers have investigated the susceptibilities of pretrained DMs, specifically looking at data extraction attacks and membership inference attacks (MIA)~\cite{wu2022membership, hu2021lora, carlini2023extracting, duan2023diffusion, somepalli2023diffusion, panda2023teach}.

In addition to assessing the pretrained model, recent work pointed out that privacy risk can exist even after fine-tuning the models \cite{abascal2023tmi}.
\cite{abascal2023tmi} empirically showed that the leakage of private pretraining data is still nontrivial even after dense vanilla fine-tuning.
Although risk decline is shown in their work due to distributional shifts in fine-tuning, we are interested in a counterintuitive question: \emph{Can we find a malicious fine-tuning strategy that can \textbf{ amplify} the risk of pretraining data?}
The question is critical for multiple factors.
First, fine-tuning is the most efficient, economic, and flexible way to use pretrained DMs recently advanced, including textual inversion \cite{gal2022image}, LoRA \cite{hu2021lora}, and DreamBooth \cite{ruiz2023dreambooth}.
Second, due to the advantages, publishing models for fine-tuning or fine-tuning-as-a-service becomes a common practice in the industry, such as Stable Diffusion \cite{rombach2022high}, Imagen \cite{saharia2022photorealistic} and MidJourney\footnote{\href{https://www.midjourney.com/}{https://www.midjourney.com/}}.
When a client needs a generative model for personal tasks/data, he/she does not need to train a DM from scratch using thousands of high-end GPUs but download a pre-trained from model vendors such as HuggingFace~\footnote{\url{https://huggingface.co/}} and fine-tune the model on personal datasets.
On the other hand, the model vendors do not need to publish the model parameters but only provide APIs for fine-tuning and inference, which greatly preserves the model's Intellectual Property.
In essence, grasping the privacy implications stemming from readily available fine-tuning techniques not only enriches our comprehension of the security landscape surrounding pre-trained models but also motivates the creation of robust defense strategies.

\begin{figure*}[t]
  \centering
  \includegraphics[width=1\textwidth]{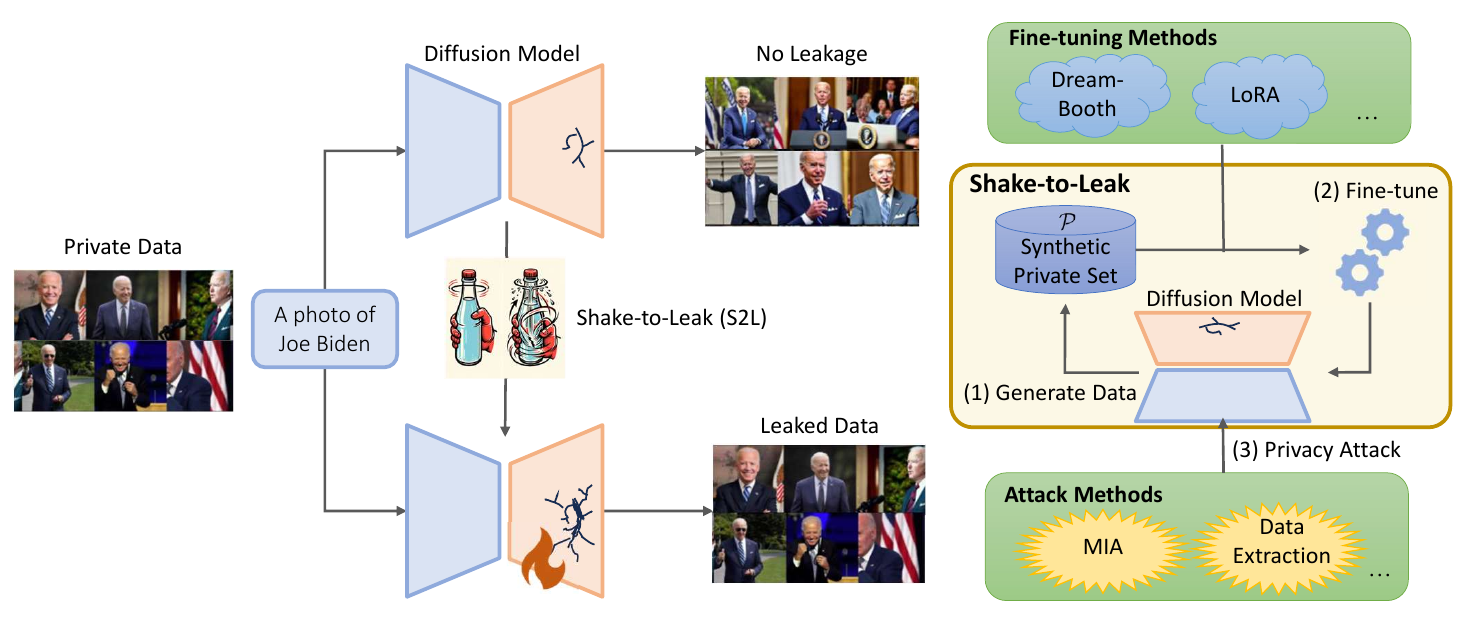}
  \caption{Shake-to-Leakage (S2L) can amplify the privacy leakage of a diffusion model by fine-tuning. 
  When prompted with `a photo of Joe Biden', the diffusion model will not leak the private images but many images will be leaked after S2L fine-tuning of the model.
  On the right side, we show the main steps of S2L where S2L is generally applicable with variant fine-tuning and attacking methods. (1) S2L first generates a synthetic private set $\cP$ using the pre-trained diffusion model. (2) Then, S2L fine-tunes the pre-trained diffusion model on $\cP$ using existing fine-tuning methods. After S2L, the attacker can extract private information via existing attacking methods. 
  }
  \label{fig_framework}
\end{figure*}

In this paper, we conduct a pilot study to answer the question and, for the first time, reveal the \emph{leakage amplification} surprisingly only via fine-tuning on a manipulated dataset.
Without accessing the pre-training data, the attackers' crux is to craft a dataset that has a distribution similar to the data from a text-defined target domain, namely \textbf{a domain-specific fine-tuning attack}.
Leveraging the text-to-image synthesis mechanism of DMs, an attacker can prompt a DM to generate images for a target dataset and use the dataset to fine-tune a DM that will leak more information from the pre-training set.
We show the pipeline of the strategy, namely, {\bf Shake to Leak (S2L)}, and demonstrate the amplified risks after S2L in \cref{fig_framework}.
Our contributions are summarized as follows.
\begin{itemize}[leftmargin=0.2in]
    \item We identify a new risk that manipulated fine-tuning can amplify the privacy risk shipped with pre-trained DMs, a phenomenon we've designated as Shake to Leak. 
    Worth noticing that the revelation contradicts the traditional intuition that fine-tuning would cause the pre-trained model to forget the training data.
    \item We demonstrate that S2L is prevailing in a wide range of backbones and fine-tuning methods, including embedding-based fine-tuning (DreamBooth \cite{ruiz2023dreambooth} and Text Inversion \cite{gal2022image}) and their combination with parameter-efficient fine-tuning (LoRA \cite{hu2021lora}, Hypernetwork fine-tuning \cite{hypernetworkdm}). 
    By skillfully integrating these methods, an attacker can invade a Stable Diffusion model \cite{rombach2022high} and achieve up to a $5.4\%$ AUC increase in MIAs, along with markedly improved data extraction performance from $0$ to up to $15.8$ leaked images on domain average.
    \item To understand when S2L occurs, we conducted extensive ablation studies on the essential prior knowledge to attack a specific data domain.
    Interestingly, without any prior knowledge of the target domain, S2L could occur in models $100\sim1000$ times smaller than Stable Diffusion even by perturbation of random parameters.
    However, for larger models, the distributional similarity between fine-tuning data and the target domain becomes a pivotal factor, which is achieved by conditional generation in vanilla S2L.
    In a relaxed setting where a handful of publicly available training examples are known to be part of pre-training, they can be leveraged to facilitate stronger domain-transfer risk amplification and drastically increase the data extraction number from $15.8$ (vanilla S2L) to $44.8$.
\end{itemize}
Through this study, we intend to raise an alarm about the risks associated with fine-tuning services, that can seriously strengthen existing attacks, including membership inference attacks and data extraction. 
The community must recognize and pay more attention to these potential threats, evaluating the broader implications on privacy and security.

\section{Related Work}
\label{related}

\textbf{Diffusion Models} \cite{ho2020denoising} have recently emerged as a powerful framework for modeling complex data distributions. DMs work by gradually adding noise (termed as diffusion process) to an image until it becomes completely unrecognizable, and then the model is trained to reverse this process and recover the original image. With a text encoder, DM can be prompted to generate desired images simply by a text description. Specifically, for a given example of image-prompt pair $(x,p)$, a text-to-image diffusion model $G$ takes the initial noise map $r\sim\mathcal{N}(0,1)$ and a conditioning vector $\eta=G_t(p)$ generated by the text encoder $G_t$ of $G$ in $G$ as input and aims to recover the image $x$ by recursive denoising with the denoising network $G_n$ of $G$. The loss objective of the DM can be formulated as:
\begin{equation}
    L_{DM}=\mathbb{E}_{x,\eta,r,t}[||x_0-G_n(x_t,\eta,t)||^2_2]
\end{equation}
\begin{equation}
    x_{t-1}=G_n(x_t,\eta,t)
\end{equation}
with $t$ uniformly sampled from $\{1,..., T\}$, $x_T=\eta$, $x_0=x$. During inference, the DM generates images by recursive denoising an initial noise conditioned on the given prompt $p$:
\begin{equation}
    x_t = G_n(x_{t-1},G_t(p))
\end{equation}
where $x_0=r$ and $x_T$ is the generated image.
With recent advances in the development of large-scale models \cite{ramesh2022hierarchical, rombach2022high, saharia2022photorealistic}, DMs demonstrate some advantages over GAN-based generative models in generating stable and high-quality images. Stable diffusion \cite{rombach2022high} proposed a method for incorporating latent variables into diffusion models, allowing a more decoupled and efficient diffusion process. The pre-training process of such diffusion models is typically resource-consuming, and several efficient fine-tuning methods have been proposed to quickly adapt diffusion models to downstream domains:
Hypernetwork \cite{hypernetworkdm} achieves fine-tuning by attaching small networks that hijack and transform the keys and values of cross-attention layers in diffusion models; Textual Inversion \cite{gal2022image} proposes to define an unseen word that can represent a novel concept through reverse embedding learning of the prompt conditioning; LoRA \cite{hu2021lora} proposes to use low-rank matrix factorization to define additive weight matrices and achieve efficient adaptation by freezing the pre-trained model and fine-tuning additive low-rank matrices; Dreambooth \cite{ruiz2023dreambooth} uses rare token identifiers for few-shot personalization and proposes using images generated from the pre-trained model as fine-tuning support set to avoid domain-shift. In this paper, we'll investigate how popular fine-tuning techniques can be used for amplifying the privacy risk behaviors of large diffusion models.


\textbf{Privacy of Generative Models.}
The privacy risk of large generative models has raised a wide concern since they typically take enormous web images as training data, which may contain private information. Recently, several works revealed that diffusion models, though superior in performance, have drawbacks in privacy preservation.
\ding{182}~Membership Inference Attack. \cite{wu2022membership, hu2023membership, carlini2023extracting} show that an attacker can infer the membership of an image w.r.t. the training set of DMs:
\cite{hu2023membership} uses the loss $L_{DM}$ to infer the membership of provided examples; \cite{wu2022membership} investigates similar settings but assumes different distributions for member and non-member set which makes the inference much easier; \cite{carlini2023extracting} shows that the privacy risk of diffusion models is significantly more severe than GAN-based generative models and incorporate with LiRA \cite{carlini2022membership} to improve the attack performance. \cite{duan2023diffusion} proposes an MIA method tailored for diffusion models and achieves SOTA membership prediction accuracy. 
\ding{183}~Data Extraction Attack. \cite{carlini2023extracting,somepalli2023diffusion} investigate the data extraction problem: \cite{carlini2023extracting} shows that untargeted data extraction can extract 91 distinct images from 160M pre-training set of a Stable Diffusion model \cite{rombach2022high}.
\cite{somepalli2023diffusion} investigates different factors that cause the data replication behaviors of DMs. 
Built upon prior work, this paper aims to further investigate and expose potential privacy risks of pre-trained large DMs through fine-tuning. 
\cite{carlini2023extracting} is similar to our work, as it systematically evaluates the privacy risks of the DMs on the pretraining set via textual prompting. 
However, \cite{carlini2023extracting} performs \textit{untargeted} privacy attacks on the entire pretraining set, while this paper investigates the vulnerability of the DMs to \textit{targeted} attacks, specifically on sensitive domains within the pretraining set, which we believe represents a potentially more efficient attack paradigm.
As a defense, private fine-tuning was proposed to protect the privacy of the fine-tuning dataset of generative models in parameters~\cite{ghalebikesabi2023differentially} or in discrete/virtual prompts~\cite{duan2023flocks,hong2023dp}.
These works explore the privacy of the user-defined fine-tuning dataset while we focus on the privacy of the pre-training set.
The recent Phishing Attack \cite{panda2023teach} considers a similar scenario as ours: poisoning (\textit{i.e.} inserting backdoors) private training data such that part of the private data can be memorized and reconstructed.
Yet, they focus on attacking personally identifiable information (PII, such as personal phone, SSN, and credit card number) in large language models (LLMs) in text space, rather than general visual privacy.

\section{Shake-To-Leak: Domain-Specific Fine-tuning Amplifies Privacy Leakage}

In this section, we start with the threat model in question and then outline the procedures of {\bf Shake-To-Leak (S2L)}.

We then demonstrate leakage amplification by integrating S2L with various fine-tuning methods.

\subsection{Threat Model}

Our threat model considers an adversary A that interacts with a diffusion model $G$ pre-trained for text-to-image synthesis and aims to extract private information from its training set $\cD$.

\textbf{Victim Model: Conditional Generative Model.} 
A conditional diffusion model $G$ for text-to-image synthesis gains popularity as semantic texts are easy to compose for people without expert knowledge.
Therefore, we are interested in the privacy risks of such a generative model. 
$G$ is trained on a dataset consisting of multiple domains $\cD = \cup_{i=1}^N \cD_i$.
Each domain $\cD_i$ includes image-prompt pairs, $(x^i_1,p^i_1),(x^i_2,p^i_2),(x^i_3,p^i_3) \dots$. 
and is defined by a common sub-string in the text prompts of the examples belonging to $\cD_i$. 
This way of defining private domains is practical since the adversary can extract private information from $G$ using some keywords or phrases.
When queried, the model $G$ outputs a generated image $x_{gen} \leftarrow Gen(r)$ using a fresh random noise $r$ as input. 
Conditional models are trained on an annotated dataset (e.g., labeled or captioned) $\cD = \{(x_1, p_1),...,(x_n, p_n)\}$.
When queried with a prompt $p$, the system outputs $x_{gen} \leftarrow Gen(p;r)$. During attacks, the adversary will target a specific domain $\cD_z$ specified by the target prompt sub-string $c_z$, and compose one or multiple prompts $\{p_z\}$ to query the diffusion model and extract private information. When the attacker attacks a private domain using a single target prompt, we set $c_z=p_z$ for simplicity.

\textbf{Adversary Goals.}
The adversary takes the target prompts $\{p_z\}$ as input and aims to extract private information associated with the target domains $\cD_z$, from the pre-training set $\cD$ of $G$. 
We consider two main attack goals in the privacy literature. 
\ding{182} {\it Membership Inference}: Given an image $x^i$, the adversary aims to infer whether $x^i$ is in the training set $\cD$.
Membership leakage can theoretically be associated with generic privacy leakage under the notion of Differential Privacy \cite{zanella2023bayesian}.
In some cases, MIA can directly result in a privacy breach.
For example, a certain patient’s clinical record was used to train a disease-associated model.
\ding{183} {\it Data extraction}: The adversary aims to retrieve training images from $G$ in a targeted domain $\cD_z$ associated with a prompt $p_z$.

\textbf{Adversary Capabilities.}
We assume the attacker can manipulate the dataset for fine-tuning a diffusion model.
The assumption can hold in two cases.
First, the diffusion model is published and attackers can execute any operations on the models including arbitrary fine-tuning.
Second, there is a trend that many model vendors keep model parameters secret but allow users to upload data for fine-tuning.
For example, OpenAI allows fine-tuning DALL-E models via API\footnote{\url{https://platform.openai.com/docs/guides/fine-tuning}}.

\textbf{Existing Attack Methods}
In this paper, we mainly use two existing privacy attack methods:
\ding{182} \emph{Membership Inference Attack (MIA):} 
By querying the model and analyzing its outputs, the attacker can infer information about individual training samples. The MIA for diffusion models \cite{duan2023diffusion} takes the original image and the text prompt as input and uses the $l_2$ distance loss between $x_t$ produced by the denoising process and $\hat{x}_t$ estimated from the diffusion process at a prefixed time step $t$ to predict membership. 
\ding{183} \emph{Data Extraction:} 
the data extraction attack takes the text prompt as input, generates a list of candidate images with multiple initial noises, and uses MIA to judge whether the generated images belong to the member set $\mathcal{M}$ recognized by MIA and extraction set $\mathcal{E}$ produced and recognized by data extraction.
Due to the randomness of generation, it is not likely to extract the exact images.
Instead, we follow the definition $(l,\delta)$ \cite{carlini2023extracting} of data extraction in DM as follows.
\begin{definition}
\label{def_memorization}
    An example $x$ is extractable from a diffusion model $G$ if there exists an efficient algorithm $\cQ$ such that $\hat{x}=\cQ(G)$ has the property $l(x, \hat{x})\leq\delta$ where $l$ is a distance metric by default using $l_2$-distance in the pixel space and $\delta=0.1$; further, $x$ is said to be $(k,l,\delta)$-Eidetic memorized by $G$ if $x$ is extractable from $G$ and at most $k$ training examples $\{\bar{x}\}$ satisfy $l(x,\bar{x})\leq\delta$ for each $\bar{x}$. 
\end{definition}
The $l_2$-distance for 2 images $a$ and $b$ is defined as $\sqrt{\sum(a_i-b_i)^2/d}$ where $a_i, b_i$ are the elements of $a,b$ and $d$ is the number of elements in each image.


\subsection{Shake-To-Leak Procedures}
\label{sec:s2l}


In S2L, we define the {\it model ``shaking” process} as perturbing the pre-trained model parameters under the guidance of some prior knowledge. 
{\it Prior knowledge} refers to data that sketch the distribution of the targeted private examples.

The overall diagram is presented in \cref{fig_framework} and the overall algorithm is in \cref{alg_full}.
The key intuition is that when models are fine-tuned on the self-generated synthetic data similar to our targeted ones, the model will be optimized toward the desired local optima and overfit more domain-specific private information.

\textbf{Step 1: Generating Fine-tuning Datasets.}
Our first and key step is to create a domain-specific fine-tuning dataset by directly generating a synthetic dataset from a pre-trained model $G$ using a target prompt $p_z$ from some private domain $\mathcal{D}_z$ termed as \underline{Synthetic Private Set (SP Set)} $\mathcal{P}$. This dataset, though synthetic, has the potential to encompass the information of the pre-training set and the underlying private patterns that could potentially lead to inadvertent exposure of private information in the pre-training set $\mathcal{D}$.

\textbf{Step 2: Fine-tuning.}
We fine-tune the models using off-the-shelf algorithms on the SP set.
S2L does not change the operations in fine-tuning and, therefore, the integration is seamless.
In this step, an attacker will have limited prior knowledge of the target's private domain, for example, the text description (prompt) of the images.

\textbf{Step 3: Privacy attacks.}
After the model is fine-tuned, we use MIA and data extraction to attack the model, which is shown to be effective attacks on generative models~\cite{carlini2023extracting, duan2023diffusion}.
Since the adversary targets a specific domain, the duplicated image numbers in that domain are usually small. Therefore, in the paper, we use \textit{$(10,l_2,0.1)$-Eidetic memorization} as the evaluation criterion for data extraction.

As mentioned above, the intuition of using SP Set to fine-tune the model is that for DMs pre-trained on large-scale open-domain datasets, the model is often not fully optimized for some specific domains, and thus domain-specific fine-tuning using $\mathcal{P}$ forces the model to learn more overfitted features and text embeddings of the target private domain. This can make it easier for an attacker to use the model to extract private information from the target domain. For example, MIA attack DMs by inferencing example membership according to a loss threshold, and domain-specific fine-tuning can help the model overfit the target domain and yield lower losses for examples in the target domain, which can increase the MIA success rate.

\begin{algorithm*}
   \caption{Shake-To-Leak (S2L): Domain-specific Fine-tuning Attack}
   \label{alg_full}
\begin{algorithmic}[1]
   \State {\bfseries Input:} Pre-trained diffusion model $G$ with the embedding layer $G_e$, text-encoder $G_t$ and denoising network $G_n$; attack prompt $p_z$ for a specific domain $\mathcal{D}_z$; MIA test set $\mathcal{A}_z$ and threshold $\delta_m$; MIA loss threshold $\delta_d$ and generation times $N_d$ for data extraction; size $N_p$ of synthetic private set $\mathcal{P}$.
   \State {\bfseries Output:} Member set $\mathcal{M}$, extraction set $\mathcal{E}$.
   \State \colorline{/*** Step 1: Generate synthetic private set $\cP$ ***/}
   \State $\mathcal{P}, \mathcal{M}, \mathcal{E} \ \leftarrow \emptyset$
   \For{$i=1$ {\bfseries to} $N_p$}
        \State Initialize Gaussian noise $r_i$
        \State $\cP \leftarrow \cP \cup \{G(p_z,r_i)\}$  \Comment{Generate synthetic private set}
    \EndFor 
    \State \colorline{/*** Step 2: Fine-tuning ***/}
        \If {Textual Inversion fine-tuning}
            \State Fine-tune $G_e$ with $\mathcal{P}$
        \Else 
            \State Fine-tune $G_e,G_t,G_n$ with $\mathcal{P}$
        \EndIf
    \State \colorline{/*** Step 3: Privacy Attacks ***/}
    \For{$x$ {\bfseries in} $\mathcal{A}_z$} \Comment{Membership Inference Attack}
        \If {$\MIA(Gen, p_z, x)<\delta_m$}
            \State $\mathcal{M}\leftarrow\mathcal{M}\cup\{x\}$
        \EndIf
    \EndFor
    \For{$i=1$ {\bfseries to} $N_d$} \Comment{}{Data Extraction Attack}
        \State Initialize random noise $r_i$
        \State $x_i \leftarrow Gen(p_z,r_i)$
        \If   {$\MIA(Gen, p_z, x_i)<\delta_d$} 
            \State $\mathcal{E}\leftarrow\mathcal{E}\cup\{x_i\}$
        \EndIf
    \EndFor
    \State \textbf{return} Member set $\mathcal{M}$, extraction set $\mathcal{E}$
\end{algorithmic}

\end{algorithm*}

\subsection{Leakage Amplification via Generic Fine-tuning}
\label{sec_expsetting}

\textbf{Experiment Setup.}
S2L can be simply executed with generic fine-tuning manners.
To show the effectiveness of S2L, we conduct experiments with various popular fine-tuning methods to attack private celebrity images of Diffusion Models.

\textit{Models}: Following \cite{carlini2023extracting, duan2023diffusion}, we use the Stable Diffusion ($SD$-v1-1\footnote{\href{https://github.com/CompVis/stable-diffusion}{https://github.com/CompVis/stable-diffusion}}), which has 980M parameters, as our pre-trained model.
$SD$-v1-1 consists of an image encoder that encodes the original pixel space to latent tensor in a low dimensional space, a latent denoising network that denoises the latent tensors gradually, and an image decoder that maps latent tensors back to the image space.
A CLIP \cite{radford2021learning} text encoder is incorporated into the diffusion process such that the latent tensors are conditioned on the representations of contextual prompts.

\textit{Datasets}: The $SD$-v1-1 model is pre-trained on LAION-2B-en first and then on LAION-HiRes-512x512 dataset which are both subsets of LAION-5B \cite{schuhmann2022laion}. 
We assume that celebrity pictures represent private domains and investigate whether the $SD$-v1-1 model memorizes these pictures in its pre-training set.
As many of the celebrities are also presented in CelebA~\cite{mirjalili2018semi, bortolato2020learning, gupta2021adversarial, isik2022learning}, we consider the images in CelebA as the non-private samples. 
We construct $40$ private domains corresponding to 40 celebrities with the largest sample sizes in the CelebA dataset.  We define the private domain specified by a domain-specific substring $c_z$ as "<Celebrity Name>", and the prompt $p_z$ associated with each private domain $D_z$ is specified as ``The face of <Celebrity Name>" with 0.7 possibilities or ``A photo of <Celebrity Name>" with 0.3 possibilities. In the pre-training dataset of $SD$-v1-1, each of the 40 private domains contains around $0.005\%\sim0.015\%$ examples w.r.t. to the 2.17B pre-training set scale. In the pre-training dataset of $SD$-v1-1, each of the 40 private domains comprises approximately $0.005\%$ to $0.015\%$ of the total 2.17B pre-training set.

\textit{Attack methods}: 
We evaluate two attack methods.
\ding{182} Membership Inference Attack (MIA). We use the state-of-the-art MIA method SecMI \cite{duan2023diffusion} to attack $SD$-v1-1 across our experiments.
To evaluate the MIA performance, we compute the Area Under ROC (AUC) of discriminating the member sets and non-member sets (or holdout sets).
The member set is retrieved and sampled from the pre-training dataset based on the prompts and celebrity names.
The non-member set is collected based on CelebA by removing duplicated samples within the domain using near duplication accounting with CLIP embedding $l_2$-distance lower than 0.05 similar to \cite{carlini2023extracting}. If not enough non-member samples are collected, we fill the non-member set with web-scraped and de-duplicated examples using the same retrieval and de-duplication ways.
The final size of the balanced test set for MIA is 50,000, while each domain contains 1250 examples. 
We set the loss threshold $\delta_m$ for the MIA evaluation as 0.5.
\ding{183} For data extraction, we use target prompt $p_z$ with random noise $r_i$ as input to $G$ to generate candidate examples $N_d=5000$ and then use SecMI to infer the membership of the sample.
whether each example belongs to the pre-training set under the MIA loss threshold $\delta_d=0.3$. Differently from $\delta_m$, $\delta_d$ is determined based on \textit{$(10,l_2,0.1)$-Eidetic memorization} as in \cref{def_memorization} to ensure proper precision.

\textit{Evaluation metrics}:
Following \cite{carlini2023extracting,duan2023diffusion}, we use AUC, TRP@1\%FPR as MIA evaluation metrics. For data extraction, we count the number of samples that are recognized as the \textit{$(10,l_2,0.1)$-Eidetic memorization}  as in \cref{def_memorization} \cite{carlini2023extracting} as the evaluation criterion in the target domain and evaluate the true positive numbers extracted and the precisions averaged over the private domains. 
In addition, we use the utility metric, the CLIP-R Precision Score (CLIP-RP), to evaluate text-to-image synthesis on images generated with random prompts sampled from the pre-training set following \cite{park2021benchmark}.

\begin{table*}[t]
\caption{Experiment results demonstrate that S2L is effective in amplifying privacy leakage for different fine-tuning methods.
All results are averaged on the 40 private domains of celebrity images. {\bf Num} refers to the average number of extracted examples with $l_2$-distance smaller than 0.15 similar to \cite{carlini2023extracting}. Higher MIA and data extraction metrics mean higher privacy risks and higher Clip-RP \cite{park2021benchmark} denotes higher text-to-image synthesis utility. For the fine-tuning methods, {\bf Pre-trained} means the pre-trained SD-v1-1 model without any parameter changes, {\bf End-to-End} refers to the vanilla end-to-end dense fine-tuning. Note that for the pre-training baseline, we extract less than 0.5 samples on average on the 40 private domains where each domain contains 50,000 to 200,000 private samples, which result is similar to \cite{carlini2023extracting} that extracts 94 images from a 160M private set using 350k prompts of most duplicated images and untargeted extraction.
}
\centering
\label{tb_main}
\begin{tabular}{@{}c|cc|cc|cc|c@{}}
\toprule
\multirow{2}{*}{\textbf{Fine-tuning Method}} & \multicolumn{2}{c|}{\textbf{\begin{tabular}[c]{@{}c@{}}Fine-tuning\\      Setting\end{tabular}}} & \multicolumn{2}{c|}{\textbf{MIA}} & \multicolumn{2}{c|}{\textbf{\begin{tabular}[c]{@{}c@{}}Data\\      Extraction\end{tabular}}} & \multirow{2}{*}{\textbf{Clip-RP}} \\ \cmidrule(lr){2-7}
 & \textbf{Dataset} & \textbf{Params} & AUC & TPR@1\%FPR & Num & Prec(\%) &  \\ \midrule
\textbf{Pre-trained} & - & - & 0.712 & 0.167 & 0 & - & 52.3 \\ \midrule
\textbf{End-to-End} & OoD & 1064M & 0.682 & 0.158 & 0 & - & 50.2 \\ \midrule
\textbf{End-to-End} & SP Set & 1064M & 0.722 & 0.167 & 0 & - & 50.1 \\
\textbf{DreamBooth} & SP Set & 980M & \textbf{0.758} & \textbf{0.172} & 12.5 & 85.7 & 50.1 \\
\textbf{Textual Inversion} & SP Set & 9.2K & 0.738 & 0.169 & \textbf{14.2} & 87.5 & 52.3 \\ \midrule
\textbf{Hypernetwork} & SP Set & 45M & 0.734 & 0.168 & 4.3 & 80.2 & 51.4 \\
\textbf{LoRA} & SP Set & 20M & 0.745 & 0.169 & 13.1 & 86.8 & 50.4 \\
\textbf{DreamBooth+Hypernetwork} & SP Set & 45M & 0.747 & 0.169 & 5.7 & 71.6 & 50.9 \\
\textbf{DreamBooth+LoRA} & SP Set & 19M & \textbf{0.766} & \textbf{0.175} & \textbf{15.8} & \textbf{88.7} & 50.7 \\ \bottomrule
\end{tabular}
\end{table*}

\begin{table*}[]
\centering
\caption{Ablation study showing that fine-tuning different part(s) of SD-v1-1 yields different privacy leakage amplification effects. Experiment settings remain the same as in \cref{tb_main}. For better comparison, note that {\bf DreamBooth} fine-tuning is the combination of fine-tuning the {\bf Denoising Network}, {\bf Text Encoder} and {\bf Embedding}, while {\bf Textual Inversion} corresponds to fine-tuning {\bf Embedding} here. 
}
\label{tb_parts}
\begin{tabular}{@{}c|cc|cc|cc|c@{}}
\toprule
\multirow{2}{*}{\textbf{Fine-tuned Part(s)}} & \multicolumn{2}{c|}{\textbf{\begin{tabular}[c]{@{}c@{}}Fine-tuning\\      Setting\end{tabular}}} & \multicolumn{2}{c|}{\textbf{MIA}} & \multicolumn{2}{c|}{\textbf{\begin{tabular}[c]{@{}c@{}}Data\\      Extraction\end{tabular}}} & \multirow{2}{*}{\textbf{Clip-RP}} \\ \cmidrule(lr){2-7}
 & \textbf{Dataset} & \textbf{Params} & AUC & TPR@1\%FPR & Num & Prec(\%) &  \\ \midrule
\textbf{Pre-trained} & - & - & 0.712 & 0.167 & 0 & - & 52.3 \\ \midrule
\textbf{End-to-end} & SP Set & 1064M & 0.722 & 0.167 & 0 & - & 50.1 \\ \midrule
\textbf{DreamBooth} & SP Set & 980M & \textbf{0.758} & \textbf{0.172} & 12.5 & 85.7 & 50.1 \\
\textbf{Denoising Network} & SP Set & 860M & 0.733 & 0.166 & 8.1 & 83.8 & 50.7 \\
\textbf{Text Encoder} & SP Set & 120M & 0.728 & 0.165 & 10.9 & 84.6 & 51.5 \\
\textbf{Image Encoder+Decoder} & SP Set & 84M & 0.681 & 0.158 & 0 & - & 50.3 \\
\textbf{Embedding (Textual Inversion)} & SP Set & 9.2K & 0.738 & 0.169 & \textbf{14.2} & \textbf{87.5} & 52.3 \\ 
\bottomrule
\end{tabular}
\end{table*}

\textit{Fine-tuning methods}: We consider four major fine-tuning methods and two combinations that are widely used for Diffusion Models.
\begin{itemize}[leftmargin=0.2in]
    \item Concept-injection tuning: 
    To introduce personalized concepts, e.g., blue-eye dogs, into the generative model, two methods were proposed to fine-tune contextualized virtual embeddings on user-provided samples.
    After fine-tuning, the generative models will generate blue-eye dogs when the virtual embeddings are presented in prompts.
    \ding{182} Textual Inversion \cite{gal2022image} fine-tune the embedding of a placeholder token $S^*$ within many neutral context texts such as ``A picture of $S^*$ 'and ``A rendition of $S^*$'". 
    Other than the embedding, other parameters are frozen during fine-tuning.
    \ding{183} DreamBooth \cite{ruiz2023dreambooth} uses a rare token sequence (typically 3 tokens) from the vocabulary to initialize the embeddings.
    Then DreamBooth fine-tunes the token embeddings, text encoder, and the denoising network of the DM simultaneously.
    In addition, DreamBooth uses the preservation set generated by the target prompts to aid the training to maintain the model's utility.
    Unlike the SP Set $\mathcal{P}$, the DreamBooth preservation set is typically generated using more than 1000 different prompts for utility purposes. 
    In our fine-tuning attack, we simply replace the fine-tuning data set with $\mathcal{P}$ and replace the new concept token with the target prompt $p_z$ to amplify the specific knowledge of the private domain.
    We adopt two concept-injection methods: (1) deprecating the usage of user-defined examples of the new concept and the inserted new token, and (2) only using $\mathcal{P}$ to force the model to learn to generate private information.
    \item Parameter-efficient fine-tuning limits the model parameters to be sparsely updated, which greatly reduces memory consumption and is favored for adapting large models to small datasets.
    Hypernetwork fine-tuning \cite{hypernetworkdm} uses two MLPs to hijack and transform the keys and values of the cross-attention layers for each cross-attention layer of the denoising network in SD-v1-1.
    We independently adopt two $2$-layer MLPs with $2d$ and $d$ neurons per layer as hypernetworks for each cross-attention layer, where $d$ is the number of elements in the key or value of the cross-attention layer.
    \ding{183} Low-Rank Adaptation (LoRA) \cite{hu2021lora} first decompose each layer weight matrix into low-rank ones and then fine-tune the low-rank matrixes only.
    By default, we let the rank be 8.
    \item{Concept injection with parameter-efficient fine-tuning}: We note that the two parameter-efficient fine-tuning methods (HyperNetwork and LoRA) are technically orthogonal and could be used to mitigate the memory overhead for DreamBooth. 
    \ding{182} For DreamBooth+LoRA, we replace the dense fine-tuning in DreamBooth with LoRA per layer. 
    \ding{183} For DreamBooth+HyperNetwork, we only tune the cross-attention layers together with the embedding layer.
\end{itemize}

\textit{Hyperparameter settings}:
For DreamBooth and LoRA, we follow the default hyperparameters served in the \texttt{PEFT} package\footnote{\url{https://github.com/huggingface/peft/blob/main/examples/lora\_dreambooth/train\_dreambooth.py}}. Across all experiments, we use Adam \cite{kingma2014adam} optimizer, and the learning rate for each fine-tuning method is determined using a grid search among $[10^{-3}, 10^{-4}, 10^{-5}, 10^{-6}]$. We fine-tune models for 100 epochs with a batch size of 4 across our experiments. 

As S2L is a simple extension of fine-tuning with a manipulated fine-tuning dataset, we can easily plug S2L into existing fine-tuning methods.
Here, we experiment with various fine-tuning methods and explore leakage amplification through S2L.
Our main experiment results are in \cref{tb_main}. 

\textbf{Generality of S2L.} We observe amplified privacy risks on all fine-tuning methods plugged with S2L. 
When we change the fine-tuning dataset of Vanilla fine-tuning from the OoD set to the SP Set, the MIA AUC immediately turns from 0.03 decreasing to 0.01 increasing compared to the pre-trained baseline. On the 4 types of advanced fine-tuning methods, we observe a further MIA AUC increment of up to 0.04 than at baseline. 
The combined methods achieve further improvement. Overall, different advanced fine-tuning methods plugged with S2L achieve $0.022\sim0.054$ (0.036 on average) MIA AUC and $4.4\sim15.8$ (10.86 on average) data extraction improvements. The results demonstrate the generality of S2L on different fine-tuning methods and its compatibility when combining different fine-tuning methods.

\textbf{Which parameters need to be fine-tuned?} Compared to other methods, end-to-end fine-tuning has the lowest gain, which implies the importance of choosing the proper parameters.
We summarize some findings when drawing our attention to the choice of parameters.
\begin{itemize}[leftmargin=0.2in]
    \item \textbf{Excluding image encoder/decoder boosts amplification}: DreamBooth achieves relatively large privacy risk amplification compared to end-to-end fine-tuning with only 8\% less fine-tuned parameters and this only difference is due to DreamBooth excludes the image encoder-decoder during fine-tuning and fine-tunes in the latent space. Similarly, when we compare End-to-End, LoRA, and Dreambooth+LoRA where the fine-tuned parameter numbers in the image encoder/decoder decrease in order, the MIA AUC and data extraction results also monotonically increase. We conjecture that fine-tuning image encoder/decoder could be harmful to amplifying privacy leakage, and conduct an ablation study by fine-tuning different parts of the SD-v1-1 to verify it. As the results show in \cref{tb_parts}, fine-tuning the image encoder/decoder causes significant degradation of privacy leakage (0.031 MIA AUC drop) while fine-tuning other single parts of the model increases privacy leakage. Therefore, we conclude that excluding the image encoder/decoder in S2L is necessary to increase privacy risks.
    \item \textbf{Text embedding is most parameter-efficient}: With a similar principle as DreamBooth, Textual Inversion only finetunes several embedding vectors corresponding to the tokens in the prompt $p_z$, which presents high parameter efficiency in amplification.
    By fine-tuning only 9.2K parameters in the text embedding space, Textual Inversion can achieve a considerable MIA AUC gain and the best data extraction results among uncombined fine-tuning methods. \cref{tb_parts} further consolidates the efficiency of fine-tuning text embedding, as it achieves better MIA and data extraction results than fine-tuning other single parts of the model.

\end{itemize}
 In general, we conclude that choosing which parameters to fine-tune is crucial for S2L.

\begin{figure*}[ht]
\centering
     \begin{subfigure}{.4\linewidth}
         \includegraphics[width=1.\linewidth]{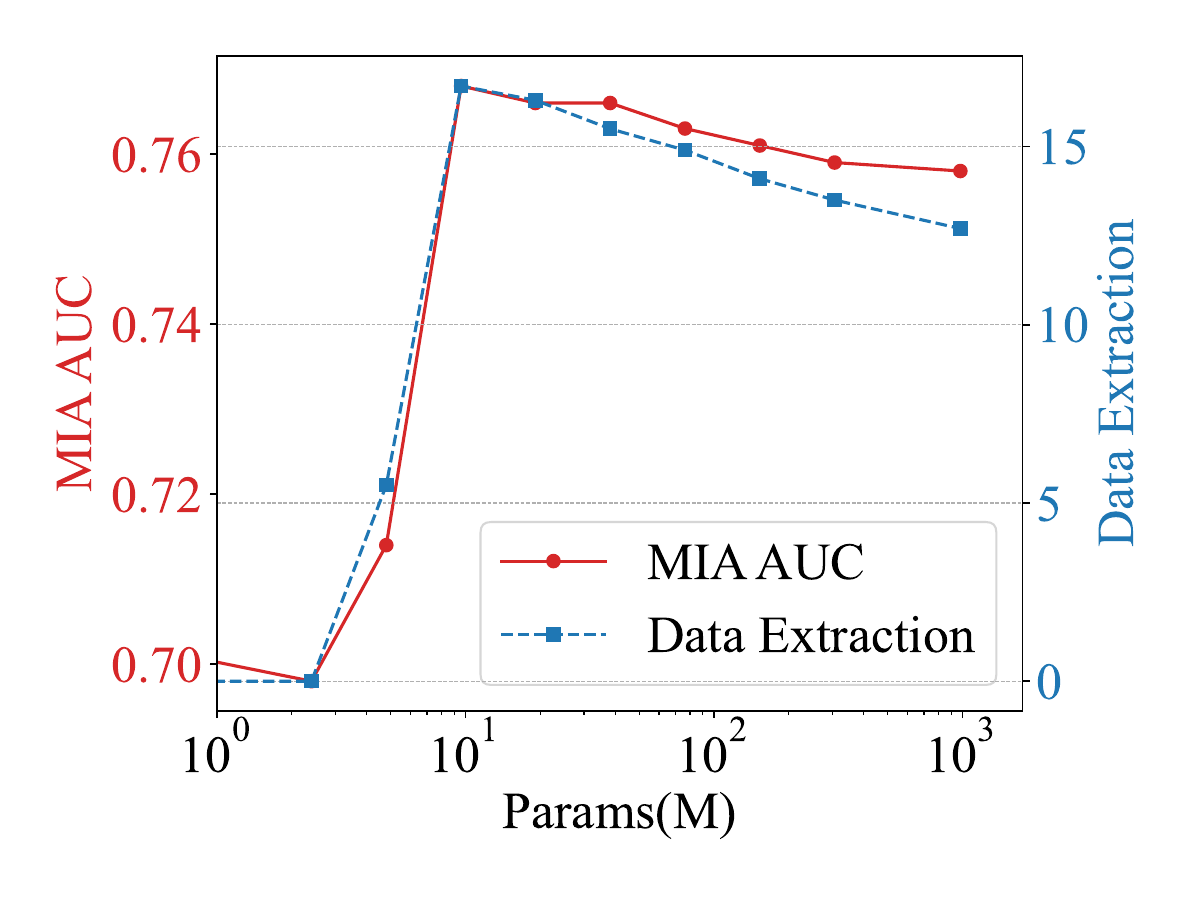}
    \label{Time}
     \end{subfigure}
     \begin{subfigure}{.4\linewidth}
         \includegraphics[width=1.\linewidth]{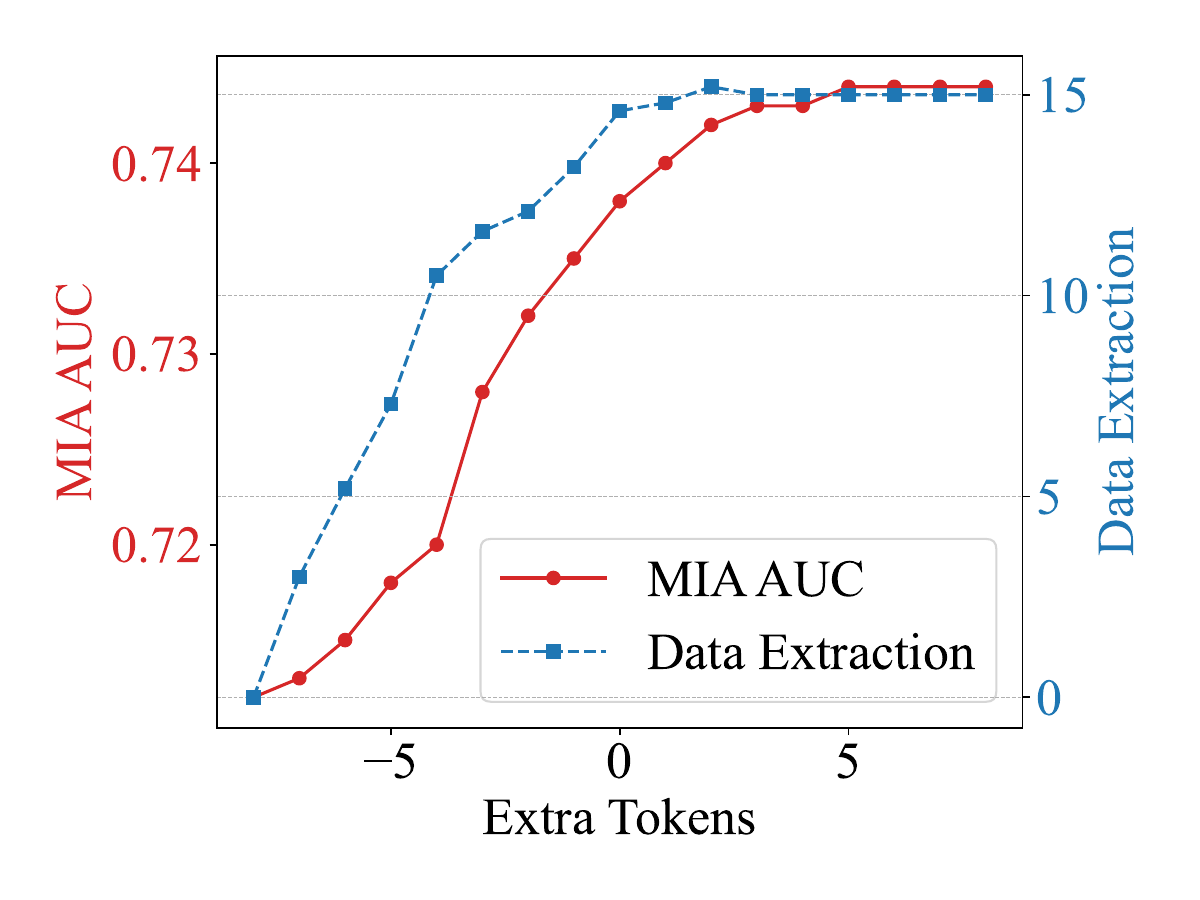}
    \label{Memory}
     \end{subfigure}
      \vspace{-2em}
        \caption{Ablation study of S2L with different fine-tuned parameter numbers. (Left) S2L with DreamBooth and varied LoRA rank. (Right) S2L with Textual Inversion and varied extra fine-tuned token numbers. Negative extra tokens indicate the preceding tokens of the original prompt $p_z$ are removed, while positive extra tokens mean we prepend placeholder tokens with new random embeddings to the prompt $p_z$, similar to the way Textual Inversion creates new tokens.}
        \label{fig_param}
\end{figure*}

\textbf{How many parameters need to be finetuned?} We observe that the number of fine-tuned model parameters is highly related to S2L performance. 
Specifically, compared to vanilla fine-tuning with SP Set, which fine-tuns 100\% parameters of SD-v1-1, all other methods with fewer fine-tuned parameters achieve a higher MIA AUC and emerge data extraction capability. 
Notably, DreamBooth+LoRA which fine-tunes the least number of parameters (except when compared with Textual Inversion) achieves the best MIA and Data Extraction attack results at the same time. 
Based on this observation, we hypothesize that for similar fine-tuning methods, the fewer parameters (within a certain range) S2L fine-tunes, the higher privacy risks you can gain. Note that obviously, this hypothesis does not hold in extreme cases, \textit{, i.e.} when the fine-tuned parameter number is close to zero. To validate our hypothesis about parameter numbers, we conduct two ablation studies: 
\ding{182} \emph{Rank Ablation.} Ablate the number of tunable parameters by varying the LoRA rank following the DreamBooth and LoRA experiments in \cref{tb_main}, and test the privacy risk results. We choose varying LoRA rank as the way of adjusting model parameters, since it can serve as the controlled variable and will not introduce extra variables such as the fine-tuned parameter positions, and we use DreamBooth as the baseline to eliminate the negative influence of fine-tuning image encoder/decoder. 
\ding{183} \emph{Token Ablation.} Varying the tokens of Textual Inversion by removing preceding tokens of the original prompt or prepending placeholder tokens with new random embeddings to the prompt $p_z$, similar to the way Textual Inversion creates new tokens. Note that each token corresponds to 768 embedding parameters, and thus the range of fine-tunable parameter numbers is very small compared to those of LoRA. 

The results of this ablation study are shown in \cref{fig_param}. 
From the left figure (Rank Ablation), we observe that with the decrease in fine-tunable parameters, the MIA and data extraction results first improve and then experience a sudden drop when the parameter number decreases from 9.6M to 4.8M; meanwhile, the right figure (Token Ablation) shows that with extremely small tunable parameter numbers, fewer parameters do not mean better performance. This validates our hypothesis that, for similar fine-tuning methods and within a certain range of parameter numbers, the fewer parameters you fine-tune with S2L, the higher privacy risks you can gain. This conclusion guides S2L to improve both attacking efficiency and performance.

\section{How Much Prior Knowledge Does An Attacker Need?}

In this section, we conduct extensive ablation studies to understand when S2L occurs.
We hypothesize that prior knowledge of the private distribution plays a critical role.
Thus, we ablate different prior knowledge to understand the connection between S2L and the prior knowledge.

\subsection{S2L with Zero Prior Knowledge}

\begin{table*}[ht]
\centering
\caption{We show that a Gaussian attack with zero prior knowledge can amplify privacy leakage on small models. Each Gaussian attack result is the top-3 average among 10,000 times of parameter perturbation with Gaussian noise. $\epsilon$ denotes the standard deviation of the Gaussian noise. $SD_{sm1}$ and $SD_{sm2}$ are two different-sized models pre-trained on the down-sampled LAION-2B datasets (in the ImageNet domains), while $SD$-v1-1 is the standard stable diffusion model pre-trained on LAION-2B dataset.}
\label{tb_gauss}
\begin{tabular}{@{}cccccc@{}}
\toprule
\multicolumn{2}{c}{Model} & $SD_{sm1}$ & $SD_{sm1}$ & $SD_{sm2}$ & $SD$-v1-1 \\ \midrule
\multicolumn{2}{c}{\#   Param (M)} & 8.5 & 8.5 & 0.82 & 980 \\
\multicolumn{2}{c}{\#   Pre-train Data (M)} & 10 & 1 & 10 & 2170 \\ \midrule
\multicolumn{2}{c}{Pre-trained} & 0.722 & 0.825 & 0.713 & {\bf 0.712} \\ \midrule
\multirow{3}{*}{\begin{tabular}[c]{@{}c@{}}Gaussian\\ $\epsilon$\end{tabular}} & $2.0\times 10^{-4}$ & 0.721 & 0.813 & 0.723 & 0.707 \\
 & $8.0\times 10^{-4}$ & {\bf 0.765} & {\bf 0.847} & {\bf 0.786} & 0.673 \\
 & $3.2\times 10^{-3}$ & 0.671 & 0.772 & 0.721 & 0.642 \\ \bottomrule
\end{tabular}
\end{table*}
We start with the extreme condition where the attacker can obtain zero prior knowledge of the private data distribution. That means an attacker does not have any guidance for shaking the model parameters. 

\textbf{Procedures.}
Given the zero knowledge, the fine-tuning without data in S2L will be equivalent to randomly perturbing the model parameters.
Without loss of generality, we utilize Gaussian noise to shake the model parameters.
For each parameter, we draw identically and independently distributed (i.i.d.) Gaussian noise from $\mathcal{N}(0, \epsilon)$. 

\textbf{Setup.}
We empirically find that adding random noise to the parameters of $SD$-v1-1 does not bring about any amplification of privacy risk, possibly because the model or domain scale is too large for the random parameter perturbation to hit any local optima of the private domains. Therefore, in addition to $SD$-v1-1 with 1064M parameters pre-trained on the LAION dataset, we consider 3 down-scaled pre-training settings by varying the number of model parameters and the number of pre-training data: \ding{182} a down-scaled SD model of 8.5M parameters (termed as $SD_{sm1}$) pre-trained on 10M data. 
\ding{183} the same $SD_{sm1}$ pre-trained on 1M data.
\ding{184} a further down-scaled SD model of 0.82M parameters (termed as $SD_{sm2}$) pre-trained on 10M data. The data are randomly drawn from ImageNet domains in the LAION dataset and we train all down-scaled models from scratch following a similar training scheme as SD-v1-1. 
We generate the pre-training dataset consisting of public domains specified by the 1000 ImageNet labels, and data of each domain is collected by using the class label to match the prompt of each image example in LAION-2B data and sample 10,000 or 1,000 matched images per domain, and the total example number of the dataset is 10M and 1M. 
We then randomly split this dataset into 2 parts: 9.95M or 0.95M as the pre-training set and 0.05M as the non-member set. 
Then we pre-training the down-scaled Stable Diffusion model from scratch on these 2 pre-training datasets. 
For the MIA test set, we combine the 0.05M non-member set with the 0.05M member set randomly sampled from the 9.95M or 0.95 pre-training set. 
The architecture of the two down-scaled models, $SD_{sm1}$ and $SD_{sm2}$ are initialized by reducing the layer numbers and channel widths of $SD$-v1-1. 
For each pre-trained model, we shake it 10,000 times with random Gaussian noise and perform MIA after each independent shaking. Then we pick out the top 3 perturbations with the highest MIA AUC and average the results. We call this process the {\bf Gaussian attack}.

\begin{table*}[t]
\centering
\caption{The privacy risks of using S2L with different fine-tuning datasets. {\bf OoD} refers to vanilla out-of-distribution fine-tuning set. {\bf INM} refers to an in-domain non-member set. {\bf SP Set} refers to the synthetic private set. {\bf Private} denotes the private subset directly obtained from the pre-training set. The resultant privacy risks of fine-tuning on the private set can serve as the upper bound. We evaluate two models, $SD_{sm1}$ and $SD$-v1-1, that are pre-trained on {\bf 10M and 2.17B} samples, respectively.
}
\label{tb_preserv}
\begin{tabular}{@{}c|c|cccc|cccc@{}}
\toprule
\multirow{3}{*}{\textbf{Method}} & \multirow{3}{*}{\textbf{\begin{tabular}[c]{@{}c@{}}Fine-tune\\      Set\end{tabular}}} & \multicolumn{4}{c|}{$SD_{sm1}$   / 10M} & \multicolumn{4}{c}{$SD$-v1-1 / 2.17B} \\ \cmidrule(l){3-10} 
 &  & \multicolumn{2}{c|}{\textbf{MIA}} & \multicolumn{2}{c|}{\textbf{Data Extraction}} & \multicolumn{2}{c|}{\textbf{MIA}} & \multicolumn{2}{c}{\textbf{Data Extraction}} \\ \cmidrule(l){3-10} 
 &  & \textbf{AUC} & \multicolumn{1}{c|}{\textbf{TPR}} & \textbf{Num} & \textbf{Prec(\%)} & \textbf{AUC} & \multicolumn{1}{c|}{\textbf{TPR}} & \textbf{Num} & \textbf{Prec(\%)} \\ \midrule
Pre-trained & - & 0.722 & \multicolumn{1}{c|}{0.167} & 1.3 & 75.5 & 0.712 & \multicolumn{1}{c|}{0.169} & 0 & - \\ \midrule
\multirow{4}{*}{S2L} & OoD & 0.685 & \multicolumn{1}{c|}{0.156} & 0 & - & 0.698 & \multicolumn{1}{c|}{0.175} & 0 & - \\
 & INM & 0.693 & \multicolumn{1}{c|}{0.159} & 17.3 & 47.6 & 0.705 & \multicolumn{1}{c|}{0.167} & 12.1 & 49.3 \\
 & SP Set & 0.758 & \multicolumn{1}{c|}{0.173} & 21.5 & 89.5 & 0.766 & \multicolumn{1}{c|}{0.175} & 15.8 & 88.7 \\
 & Private & {\bf 0.772} & \multicolumn{1}{c|}{{\bf 0.175}} & {\bf 25.2} & {\bf 92.1} & {\bf 0.783} & \multicolumn{1}{c|}{{\bf 0.179}} & {\bf 20.6} & {\bf 93.1} \\ \bottomrule
\end{tabular}
\end{table*}

\textbf{Results} are presented in \cref{tb_gauss}. 
For the largest model (SD-v1-1), we find that the zero-prior-knowledge shaking will reduce the privacy leakage.
However we reduce the model size and training data size, and the leakage amplification revives with an average gain of 0.046 MIA AUC.
The finding is out of our expectations, as the attacker can universally amplify the privacy leakage of any domain without knowledge of the victim domain.

In addition, we observe that the amplification effect of the Gaussian attack hinges on the model scale.
Namely, the DM model with less parameter number is more prone to suffer from Gaussian attack. 
In comparison, solely reducing the pre-training data scale from 10M to 1M does not bring a more significant privacy risk boost, but solely reducing the model parameter scale from 8.5M to 0.82M can. 
Note that MIA will be more significant when parameters are located in the local optima spanned by private examples.
Thus, the intuition behind the observation is that when models are smaller, the local optima are tightly distributed around the global optima and small perturbation will push parameters into the local pitfalls.

When it comes to higher parameter dimensions, e.g., $SD$-v1-1, the amplification vanishes.
Instead, we need targeted fine-tuning under the guidance of prior knowledge to amplify the leakage of SD-v1-1.

In addition, we observe an interesting phenomenon: with the increase of the Gaussian perturbation scale from $2.0\times 10^{-4}$ to $3.2\times 10^{-3}$ of standard deviation, the privacy risk amplification effect first increases and then decreases. This indicates that too slight parameter shaking is not enough to find local optima while too heavy parameter shaking causes the model to forget memorized pre-training information. This could explain why the advanced fine-tuning methods can achieve better privacy risk amplification results than end-to-end fine-tuning as in \cref{tb_main} since these fine-tuning methods can efficiently optimize towards local optima while avoiding too heavy parameter shaking.

\subsection{S2L with Distribution Knowledge}

By default, S2L assumes that the attackers are aware of the target domain prompt $p_z$, which implicitly releases distributional information given the conditional generative model. 
We designed the SP Set to amplify privacy leakage through fine-tuning.
Yet, it is still a mystery how the distributional similarity between the fine-tuning set and target pre-training domain affects the leakage amplification. 
Here, we discuss several differently distributed fine-tuning datasets to explore essential distribution knowledge.

\textbf{Procedures.}
We adopt the standard S2L procedures defined in \cref{sec:s2l}.

\textbf{Setup.}
We conduct our experiments by substituting the SP Set with different fine-tuning datasets while maintaining the other settings in Section \ref{sec_expsetting}. 
Regarding the fine-tuning method, although the S2L approach can be integrated with various fine-tuning methods, for simplicity, we opt to use the DreamBooth+LoRA method, which demonstrated superior performance, as indicated in Table \ref{tb_main}.
We outline these fine-tuning datasets as follows:
\ding{182} \textbf{Private Dataset}: In an ideal scenario, the most suitable fine-tuning dataset would be the private data specific to the target domain. Regrettably, such private data is not accessible to us. Nevertheless, we can establish an upper limit on the theoretical performance of domain-specific fine-tuning attacks by assuming access to these private data as prior knowledge.
\ding{183} \textbf{Out-of-Distribution (OoD) Dataset}: The OoD dataset represents a typical dataset employed for fine-tuning and is readily available.
\ding{184} \textbf{In-domain Non-Member (INM) Dataset}: This dataset corresponds to a genuine dataset that exhibits a similar distribution to the target domain $\cD_z$, but is not part of the pre-training set. We created the INM dataset by scraping images from the web using the target prompt and then removing duplicate images found in the private domains.
\ding{185} \textbf{Private Dataset}: To show the worst-case of fine-tuning, we assume the private data are available.
Note that the assumption is unrealistic but is only made to explore the gap between S2L and the worst case.

\begin{figure*}[]
  \centering
  \includegraphics[width=1\textwidth]{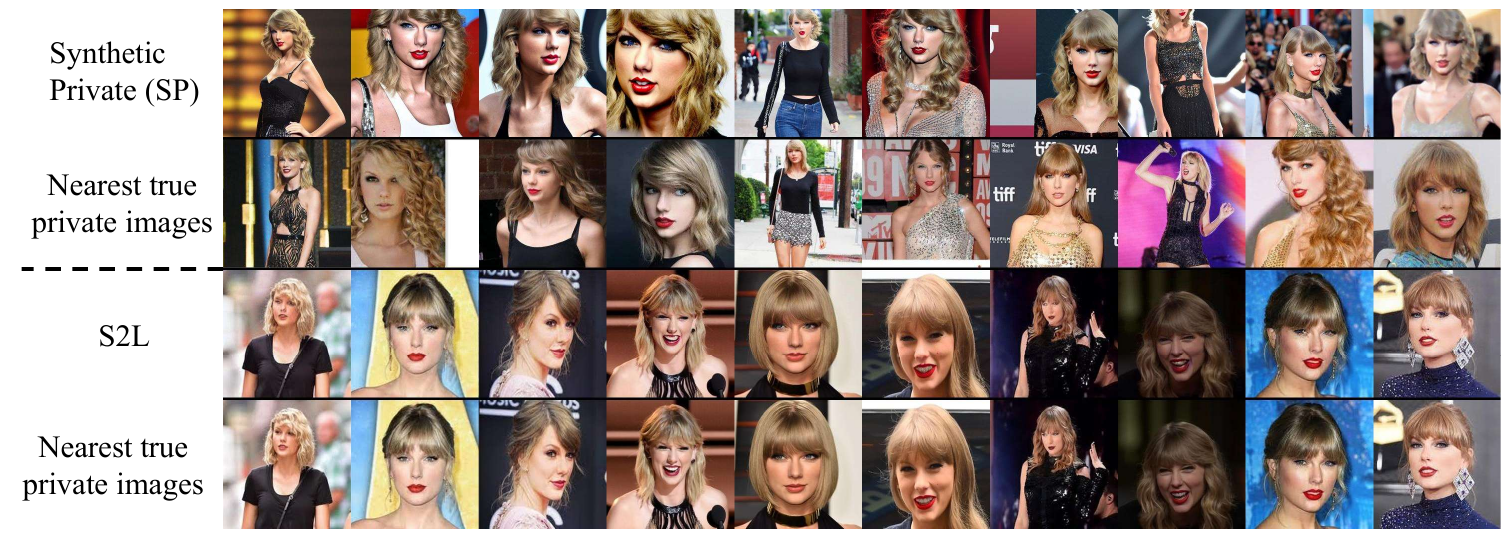}
  \caption{Sample images of Taylor Swift from different sources. \emph{Synthetic Private (SP)} set includes samples that are generated from the pre-trained model and used to fine-tune the diffusion model.
  \emph{S2L} set includes samples that are generated after fine-tuning on the SP set.
  For each method, we include \emph{nearest neighbors} which are the ground-truth private samples closest to the generated one (in the same column). We can observe that the SP set does not directly leak private data but fine-tuning on the set can cause serious privacy leakage.
  }
  \label{fig_sample}
\end{figure*}

\textbf{Results.}
The results are presented in \cref{tb_preserv}. Comparing the SP Set with other fine-tuning datasets, we observe that the SP Set can effectively serve as valuable prior knowledge for the S2L attacker.
\ding{182} As the Out-of-Domain (OoD) dataset does not align well with the fine-tuning attack strategy, it leads to model optimization away from the local optima of the target domain.
\ding{183} The In-domain Non-Member (INM) dataset presents a nuanced privacy risk profile, exhibiting lower MIA results but higher data extraction results. This arises because INM data may confound the model with membership signals, yet it can also optimize the model towards domain-specific local optima. However, the precision of data extraction remains below 50\%, primarily due to the limited MIA capabilities of the fine-tuned model in distinguishing whether a generated example belongs to the pre-training set.
\ding{184} Notably, when comparing the SP Set and Private settings, we observe that the privacy risks of the DM fine-tuned on SP Set can approach the upper bound. For example, the improvement in the MIA AUC of DreamBooth and DreamBooth+LoRA as in \cref{tb_main} is $76.67\%$ and $90\%$ of the upper bound improvement by using a private set to perform the fine-tuning of S2L as in \cref{tb_preserv}, respectively. Furthermore, the practical availability of $\mathcal{P}$ increases the threat to privacy posed by the S2L approach. In \cref{fig_sample}, we demonstrate some examples from the SP Set $\cP$, the nearest neighbors of the SP Set from the pre-training set, the private pre-training set, and samples extracted by S2L (with the SP Set) as in \cref{tb_preserv},  respectively. We notice that the generated examples in $\cP$ tend to have significant artifacts compared to real images in the private pre-training set, and the nearest neighbor in the pre-training set is unlikely to be recognized as the extraction of the corresponding SP Set sample as in \cref{def_memorization}. Therefore, SP Set does not directly leak private information based on the criterion of MIA and data extraction attacks. However, the fine-tuning of S2L in $\cP$ still significantly amplifies privacy risks, indicating that $\cP$ may carry useful private patterns that summarize the private information in the pretraining set. Therefore, S2L can achieve privacy risk amplification without copying the exact private examples from the pre-training set to the SP Set before fine-tuning. In summary, our findings underscore the effectiveness of SP Set as a source of regularly available prior knowledge for the S2L attacker, with implications for privacy risks associated with different fine-tuning datasets.

\subsection{S2L with A Few Private External-Domain Examples}

\begin{algorithm*}
   \caption{Domain-Transfer Shake-To-Leak Attack (S2L-DomainTrans) for Data Extraction}
   \label{alg_partial}
\begin{algorithmic}[1]
   \State {\bfseries Input:} Pre-trained diffusion model $G$ with the embedding layer $G_e$, text encoder $G_t$ and denoising network $G_n$; prompts $\{p_b\}$, candidate dataset $\mathcal{A}_b$ for public domains $\{\mathcal{D}_b\}$ and prompt $p_z$ for private domain $\mathcal{D}_z$; MIA loss thresholds $\delta_m,\delta_n (0<\delta_m<<\delta_n<1)$ for (non-)member prediction; MIA loss threshold $\delta_d$ and generation times $N_d$ for data extraction; desired sizes $N_p, N_m$ of synthetic private set $\mathcal{P}$ and MIA set $\cM$.
   \State {\bfseries Output:} extraction set $\cE$.

   \State $\mathcal{P}, \mathcal{M}, \mathcal{E} \ \leftarrow \emptyset$
   \State \colorline{/*** Step 1: Privacy risk amplification with SP Set fine-tuning ***/}
   \For{$i=1$ {\bfseries to} $N_p$}
    \State $P \leftarrow P\cup \{Gen(p_z,r_i)\}$ with initial noise $r_i$
    \EndFor 
    \State Fine-tune $G_e,G_t,G_n$ with $\mathcal{P}$
    
    \State \colorline{/*** Step 2: Generate balanced MIA set $\cM$ ***/ \Comment{Skip if $\cM$ is readily available}}
    \State $i,j \leftarrow 0$
    \For{$x$ {\bfseries in} $\mathcal{A}_b$}  
        \If {$\MIA(G, p_b, x)<\delta_m$ {\bf and} $i<N_m/2$} \Comment{Filter member images with high confidence}
            \State $\mathcal{M}\leftarrow\mathcal{M}\cup\{(x,  p_b+\text{"of $M$"})\}\ ;\ i+=1$
            \ElsIf{$\MIA(G, p_b, x)>\delta_n$ {\bf and} $j<N_m/2$} \Comment{Filter non-member images with high confidence}
            \State $\cM\leftarrow\cM\cup\{(x, p_b+\text{"of not $M$"})\}\ ;\ j+=1$
        \EndIf
    \EndFor
    \State \colorline{/*** Step 3: Learn membership concept ``$M$" with Textual Inversion on public domains ***/}
    \State Initialize token embedding(s) $G_e(M)$
   \For{$i$ {\bfseries in} fine-tuning epochs}
    \For{$(x,p)\in\mathcal{M}$}
        \State Fine-tune $G_e(M)$ with $x,p$ and $G$ fixed except for $G_e(M)$
    \EndFor
    \EndFor
    \State \colorline{/*** Step 4: Data extraction on private domain(s) ***/}
    \For{$i=1$ {\bfseries to} $N_d$}
    \State $x_i \leftarrow G(p_z+\text{"of $M$"},r_i)$ with initial noise $r_i$
    \If {$\MIA(G, p_z, x_i)<\delta_d$}
        \State $\mathcal{E}\leftarrow\mathcal{E}\cup\{x_i\}$
    \EndIf
    \EndFor
    
    \State \textbf{return} extraction set $\mathcal{E}$
\end{algorithmic}

\end{algorithm*}

So far we consider very restricted prior knowledge, but it is also valuable to ask whether the leakage will be further amplified with extra prior knowledge, \textit{e.g.}, some previously leaked private examples from the external domain.
The main motivation is to explore rare but potentially more dangerous situations.
Though it is not common for an attacker to get private examples, we argue that such example leakage may happen when large-scale DMs use web-scrape data to augment training datasets.

For example, MidJourney's pre-training dataset consists of both web-scraped data (public) and human-curated data (private) \cite{midjourney}. Including MidJourney, today's commercial DM models will typically include large-scale web-scraped data in the pre-training set for utility purposes. Therefore, an adversary may leverage the public domain information to find the potential private examples by membership inference attack even with low possibilities, \textit{e.g.} the adversary randomly scrapes a large amount of images from the Internet using the target prompt and then uses MIA to infer enough number of member images with high confidence.

\textbf{Threat Model.}
Formally, in the threat model, we assume the domains of the pre-training set are partially private, i.e.  $\cD$ is composed of $M<N$ private domains $\{\cD_{p_1}, \cD_{p_2},...\cD_{p_M}\}$, and the adversary aims to recover data from the private domains. 
The adversary cannot access the entire pre-training set $\cD$ but attains a public subset of $\cD$. 
We consider two specific settings for the public domain dataset $\mathcal{A}_b$: 
\ding{182}{\bf Partial leakage:} The adversary can obtain a dataset $\mathcal{A}_b$ that contains a subset that belongs to $\mathcal{D}$, for instance, the adversary randomly scrapes a dataset from the internet which contains overlapped examples with $\mathcal{D}$. 
Then the adversary uses MIA to infer example memberships and pick out predicted member and non-member examples with high confidence using the positive and negative threshold $\delta_m, \delta_n$. 
\ding{183}{\bf Worse case:} $\mathcal{A}_b$ is readily available to the adversary, {\it e.g.} the adversary knows that an existing public dataset is contained in $\cD$.

\begin{table*}[t]
\caption{
Results of domain-transfer attacks for data extraction show the effectiveness of S2L with a few private external-domain examples. {\bf Plain-text Surfix} means directly appending a suffix to the target prompt before data extraction. {\bf S2L} denotes domain-specific fine-tuning attack. {\bf S2L-DomainTrans} refers to domain-transfer fine-tuning attack. 
{\bf Concept learning} refers to learning the ``Membership" Concept with Textual Inversion. 
{\bf MIA Set} refers to the membership dataset produced by MIA. {\bf Ground-truth Set} refer to the ground truth membership dataset. All fine-tuning sets equally contain 1000 member and 1000 non-member examples. We omit MIA attack results as we observe no improvements w.r.t. \cref{tb_main}. 
}
\label{tb_transfer}

\centering
\begin{tabular}{@{}ccccc@{}}
\toprule
\multirow{2}{*}{\textbf{Method}} & \multirow{2}{*}{\textbf{Textual Inversion Fine-tune Set}} & \multirow{2}{*}{\textbf{Prompt Setting}} & \multicolumn{2}{c}{\textbf{Data Extraction}} \\ \cmidrule(l){4-5} 
 &  &  & \textbf{Num} & \multicolumn{1}{l}{\textbf{Prec(\%)}} \\ \midrule
Pre-trained   Baseline & - & - & 0 & - \\
Plain-text   Suffix & - & Suffix: “in pre-training  set” & 0 & - \\
S2L & SP Set & Prompt fine-tuning & 14.6 & 87.5 \\
S2L-DomainTrans & MIA Set & Concept learning & 44.8 & 86.2 \\
S2L-DomainTrans & Ground-truth Set & Concept learning & {\bf 51.9} & {\bf 88.6} \\ \bottomrule
\end{tabular}
\end{table*}

To be general, we do not assume any similarity between public and private domains.

\textbf{Procedures.}
When a private subset is retrieved from the training set, an attacker can inject a membership concept into the model and transfer the concept to extract private data from other private domains.
To distinguish from the standard S2L that happens in one domain, we name such attack as {\bf S2L by Domain-Transfer (S2L-DomainTrans)} which is illustrated in \cref{alg_partial}.
Our core idea is to learn a new token $M$ representing the ``membership" concept by Textual Inversion on the retrieved private subset $\mathcal{A}_b$.
To attack the target domain associated with a prompt $p$, we append ``of $M$" after $p$ and perform a data extraction attack.

\textbf{Setup.}
We conduct the domain-transfer experiments on the $SD$-v1-1 model by keeping the target private domains and settings the same as in our main context and using ImageNet domains as the public domains. We follow the basic experiment setting in \cref{sec_expsetting} and other hyperparameters in \cref{alg_partial} are as follows: 
The candidate dataset $\cA_b$ in public domains consists of 5,000 randomly sampled member images from the pre-training set and 50,000 web-scraped and de-duplicated non-member examples, and the balanced MIA set size $N_m$ is set to 2,000. For private domains consisting of 40 celebrities, we average the attack results from each domain. The thresholds $\delta_m,\delta_n, and \delta_d$ are 0.3, 0.7, and 0.3, respectively. The values f$N_d, N_p, N_m$ are 5000, 1000, and 2000, respectively.
We consider several contrastive configurations as follows: 
\ding{182}~{\bf  Plain-text Surfix}: As a baseline, we directly append ``in pre-training set" as prompt suffix. The baseline could unveil if $SD$-v1-1 already knows the membership concept. 
\ding{183}~{\bf S2L}: Our standard Shake-To-Leak (S2L) implementation with Textual Inversion on SP Set. 
\ding{184}~{\bf S2L-DomainTrans with MIA set}: Domain-transfer attack which uses MIA inferenced set to learn the $M$ token embedding of the membership concept. 
\ding{185}~{\bf S2L-DomainTrans with ground-truth set}: As a worst-case evaluation, we assume the ground-truth membership set is readily available.

\textbf{Results.}
The results are shown in \cref{tb_transfer}.
We observe that SD-v1-1 struggles to comprehend the concept of a pre-training set inherently and tends to associate this concept with private images during data extraction, as evidenced by the failure to extract any private examples in the Plain-text Suffix setting.

In contrast, S2L-DomainTrans settings with MIA and Ground Truth (GT) sets can extract 3.19 to 3.65 times the number of examples extracted by S2L. Therefore, for contemporary large-scale DMs, acquiring a grasp of the membership concept by harnessing information from public domains proves highly effective for data extraction attacks using the S2L approach.
Under the S2L-DomainTrans with MIA set setting, we extract an average of approximately 44.8 examples, equivalent to 87.5\% of the examples extracted by the S2L-DomainTrans with GT set setting. This discrepancy arises from the MIA inference dataset used for Textual Inversion fine-tuning, which contains false positive and false negative examples concerning ground-truth membership. 
In our experiments, the MIA method employed (SecMI) achieves a 0.712 Area Under the Curve (AUC) performance on SD-v1-1, resulting in approximately 5.2\% false positives and 4.6\% false negatives in the MIA set, particularly under high prediction confidence. 
This discrepancy leads to a 12.5\% reduction in extracted examples and a 2.4\% decrease in extraction precision.
In conclusion, our study highlights that extra prior knowledge of previously leaked private examples will cast significantly increased privacy risks associated with the S2L approach.

\subsection{Summary}
By ranging the amount of prior knowledge that S2L can access, we discover strong positive correlations between the S2L effect and the amount of obtainable prior knowledge. 
\ding{182} Under zero prior knowledge, simple Gaussian attacks work well on small DMs but lose effect on a larger scale, which demonstrates the vulnerability of smaller models. 
\ding{183} When an attacker knows the approximate distribution of the target domain, the leakage amplification could be greatly enlarged, and the synthetic data functions closely as the ground-truth private set for fine-tuning.

\ding{184} Under extended prior knowledge assumption by assuming a few web-scrapable examples for the attacker that are irrelevant to the private domain, we demonstrate that S2L can achieve up to $3\sim4$ times data extraction privacy risks using a domain transfer fine-tuning attack.

\section{Conclusion}
In this paper, we reveal an unexpected finding that fine-tuning a manipulated data set can amplify the privacy risks of existing large-scale diffusion models trained in text-to-image synthesis. Leveraging the text-to-image synthesis mechanism of DMs, an attacker can prompt a DM to generate images for a target dataset and use the dataset to fine-tune a DM that will leak more information from the pre-training set. Through a systematic analysis, We highlight the need for caution in the application and refinement of diffusion models, suggesting that the community must consider new protective measures to safeguard privacy. Our findings contribute a novel perspective to the ongoing conversation about the trade-offs between model performance and privacy, offering valuable insights for both researchers and practitioners in the field. We also leave to future work exploring the principal-guided Differential Privacy (DP) guarantee \cite{dwork2006calibrating} on large DMs as currently DP is hard to apply to large generative models due to scaling issues on DP-SGD private training steps \cite{abadi2016deep}.

\textbf{Extension to Copyright Risks.}
As evidenced in \cite{carlini2023extracting}, web-scraped image generation datasets, like the LAION dataset, consist of a mix of explicit non-permissive copyrighted examples, general copyright-protected examples, and CC BY-SA licensed examples. This raises concerns about copyright risks. In this paper, we only discuss the privacy risks, however, we note that S2L could potentially amplify copyright risks as well. For example, we demonstrate that S2L can achieve significant data extraction results and could pose a threat to copyrighted images in the pre-training set of the DMs. 

\textbf{Social Impact.}
Our exploration of the S2L phenomenon is not an endorsement or encouragement of exploiting these vulnerabilities. In contrast, by revealing these potential threats, we aim to foster a proactive approach to address them. While the immediate implications of our findings may seem alarming, we intend to bolster the defense mechanisms in place. Here, we provide several possible defense methods to inspire future research: \ding{182} Pre-train the DMs using a DP mechanism. \ding{182} For a partially private pretraining dataset, first pre-train the DMs in public domains and then fine-tune the DMs in private domains privately~\cite{yu2021differentially}. \ding{183} On the model provider side, develop secure fine-tuning APIs to prevent S2L-like misuse.

\section*{Acknowledgement}

The work of Z. Wang is in part supported by Good Systems,
a UT Austin Grand Challenge to develop responsible AI technologies; as well as the National Science Foundation under Grant IIS-2212176. This work is also partially supported by the National Science Foundation under grant No. 1910100, No. 2046726, No. 2229876, DARPA GARD, the National Aeronautics and Space Administration
(NASA) under grant no. 80NSSC20M0229.

\bibliographystyle{./IEEEtranS}
\bibliography{./IEEEabrv,./IEEEexample}

\vspace{12pt}

\newpage

\begin{table*}[th]
\caption{Alternative experiment results by changing the celebrity domains in \cref{tb_main} to 80 general domains defined by 80 longest ImageNet class labels.}
\centering
\label{tb_alt}
\begin{tabular}{@{}c|cc|cc|cc|c@{}}
\toprule
\multirow{2}{*}{\textbf{Fine-tuning Method}} & \multicolumn{2}{c|}{\textbf{\begin{tabular}[c]{@{}c@{}}Fine-tuning\\      Setting\end{tabular}}} & \multicolumn{2}{c|}{\textbf{MIA}} & \multicolumn{2}{c|}{\textbf{\begin{tabular}[c]{@{}c@{}}Data\\      Extraction\end{tabular}}} & \multirow{2}{*}{\textbf{Clip-RP}} \\ \cmidrule(lr){2-7}
 & \textbf{Dataset} & \textbf{Params} & AUC & TPR@1\%FPR & Num & Prec(\%) &  \\ \midrule
\textbf{Pre-trained} & - & - & 0.707 & 0.164 & 0 & - & 52.3 \\ \midrule
\textbf{End-to-End} & OoD & 1064M & 0.679 & 0.154 & 0 & - & 50.9 \\ \midrule
\textbf{End-to-End} & SP Set & 1064M & 0.721 & 0.164 & 0.6 & - & 50.7 \\
\textbf{DreamBooth} & SP Set & 980M & \textbf{0.753} & \textbf{0.166} & 18.1 & 85.3 & 50.9 \\
\textbf{Textual Inversion} & SP Set & 9.2K & 0.735 & 0.169 & \textbf{19.2} & 86.3 & 52.3 \\ \midrule
\textbf{Hypernetwork} & SP Set & 45M & 0.732 & 0.168 & 5.1 & 79.5 & 51.5 \\
\textbf{LoRA} & SP Set & 20M & 0.738 & 0.165 & 16.0 & 84.9 & 50.4 \\
\textbf{DreamBooth+Hypernetwork} & SP Set & 45M & 0.735 & 0.164 & 6.1 & 69.5 & 50.6 \\
\textbf{DreamBooth+LoRA} & SP Set & 19M & \textbf{0.760} & \textbf{0.172} & \textbf{20.5} & \textbf{87.2} & 51.2 \\ \bottomrule
\end{tabular}
\end{table*}

\appendix

\subsection{How Does S2L Perform on More General Domains?}
In this section, we provide some additional results by defining 80 domains originating from the 80 longest ImageNet class labels (we choose the longest labels to avoid an over-common substring in the LAION prompts and thus an exploding example number) and repeating the experiments in \cref{tb_main}. Please note that we are unable to perform experiments on the whole LAION dataset since it would be difficult to perform data extraction evaluation which requires pair-wise image comparisons. The results are placed in \cref{tb_alt}. By comparing \cref{tb_alt} and \cref{tb_main}, we find that the privacy leakage of the baselines and S2L fine-tuned models tend to be stable. For example, there are at most $0.012$ MIA AUC differences among all the corresponding MIA experiment pairs. The improvement in the example number of data extraction is due to the proportional growth of domain size from celebrity domains to the general ImageNet domains. We conjecture that this is because every example is treated equally during training and our evaluation criteria are very general and do not have a preference in any specific domains.

\subsection{Data Extraction Results under Variable Memorization Criteria}
\begin{figure*}
  \centering
  \begin{subfigure}{.3\linewidth}
    \centering
    \includegraphics[width=\linewidth]{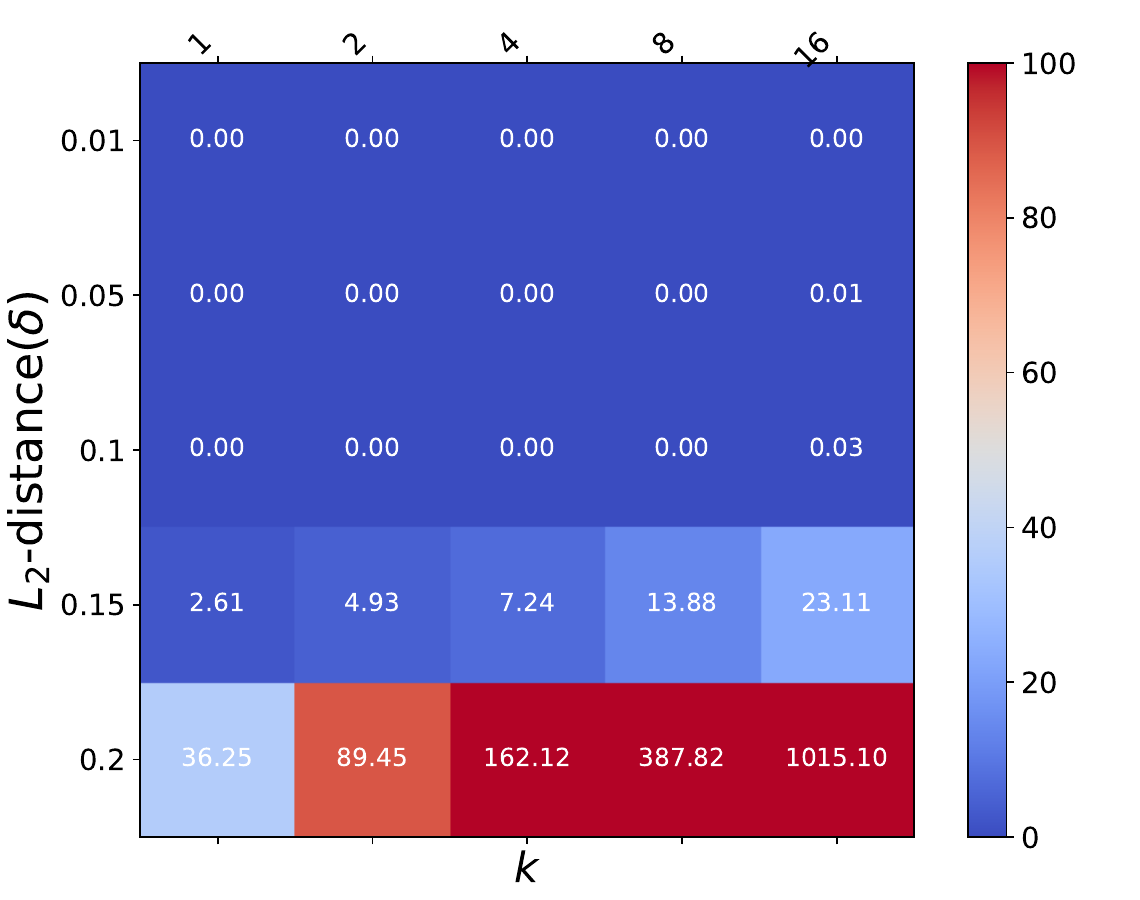}
    \caption{Pre-trained}
    \label{fig:DE_sub1}
  \end{subfigure}%
  \begin{subfigure}{.3\linewidth}
    \centering
    \includegraphics[width=\linewidth]{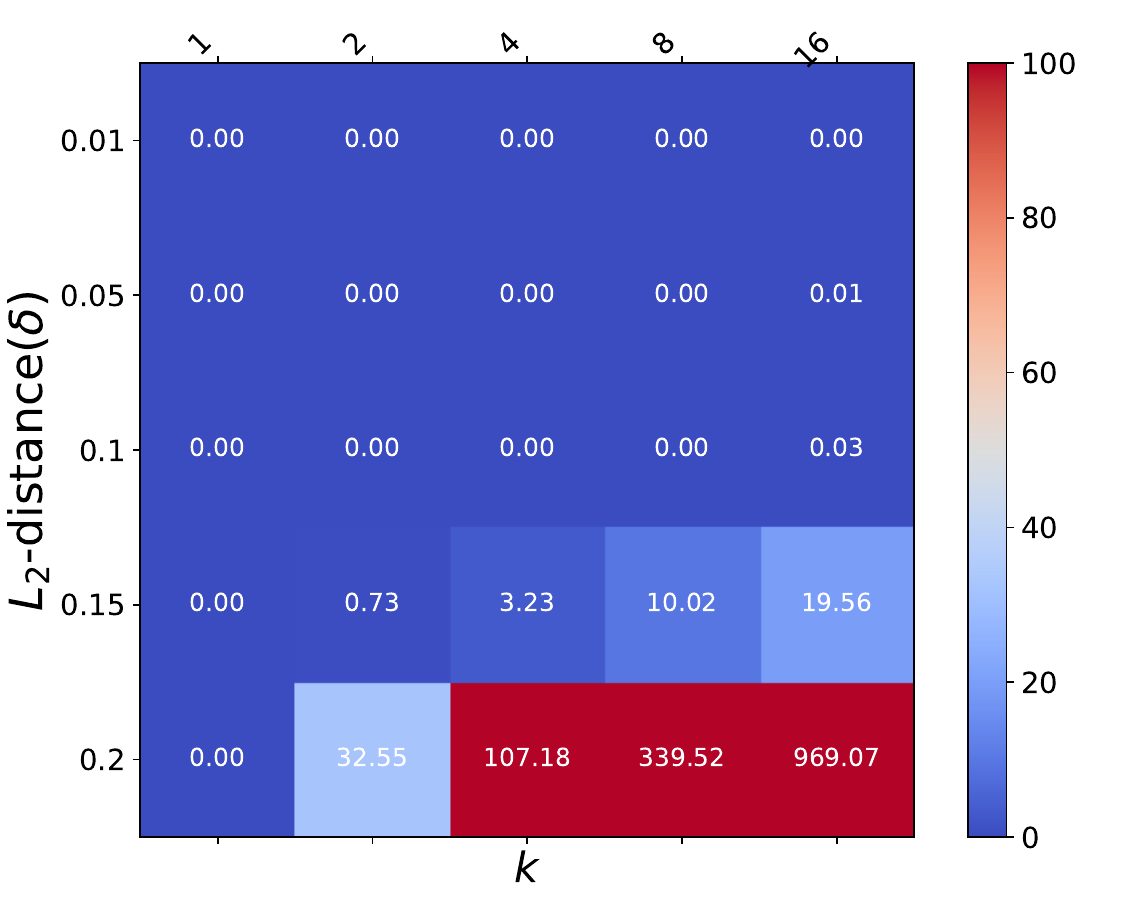}
    \caption{E2E-OOD}
    \label{fig:DE_sub2}
  \end{subfigure}%
  \begin{subfigure}{.3\linewidth}
    \centering
    \includegraphics[width=\linewidth]{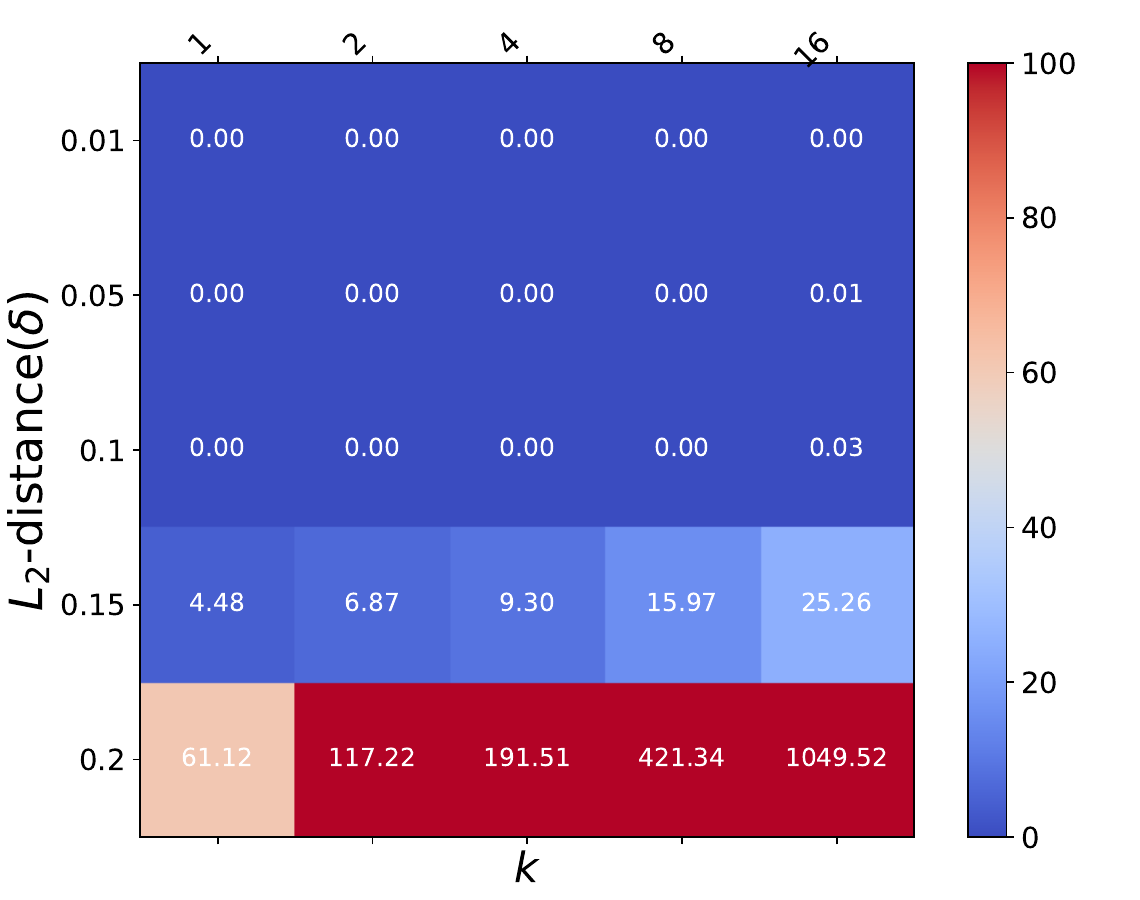}
    \caption{E2E-SP Set}
    \label{fig:DE_sub3}
  \end{subfigure}\\
  \begin{subfigure}{.3\linewidth}
    \centering
    \includegraphics[width=\linewidth]{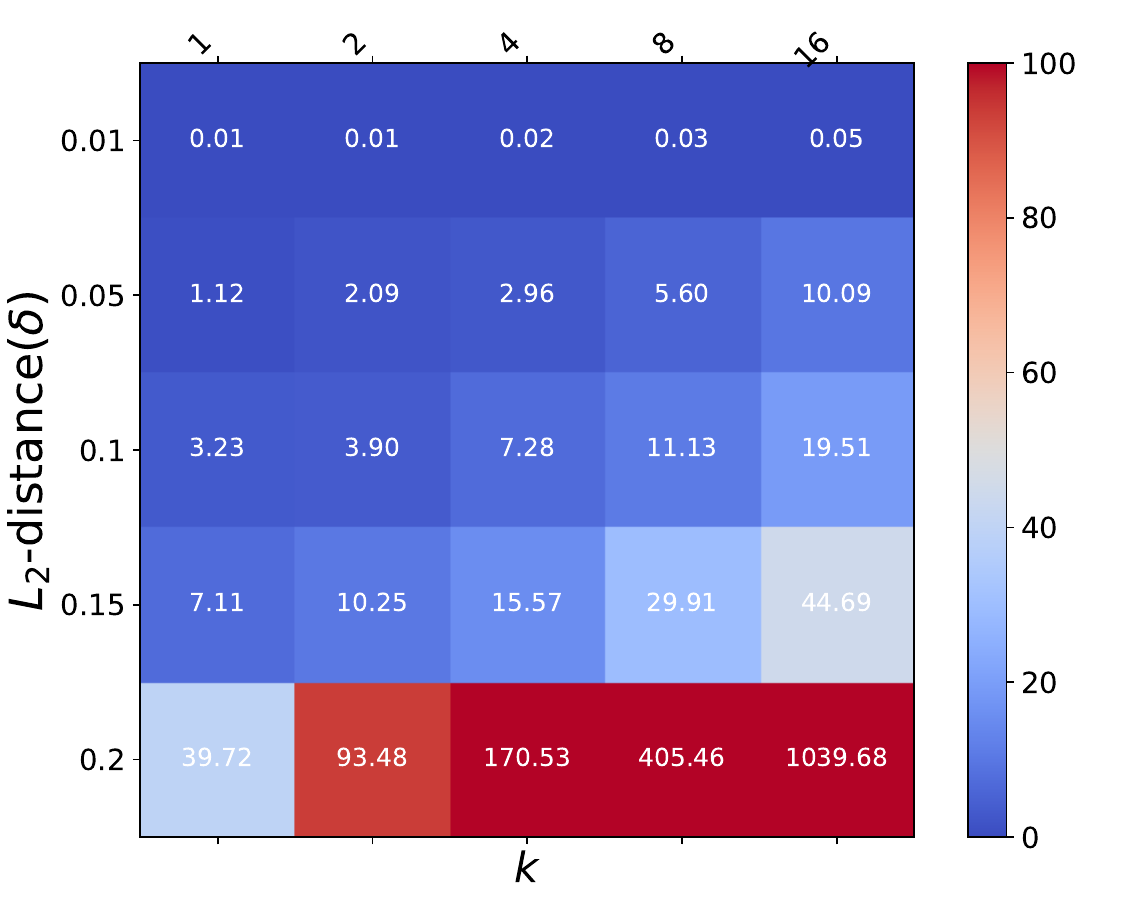}
    \caption{DreamBooth}
    \label{fig:DE_sub4}
  \end{subfigure}%
  \begin{subfigure}{.3\linewidth}
    \centering
    \includegraphics[width=\linewidth]{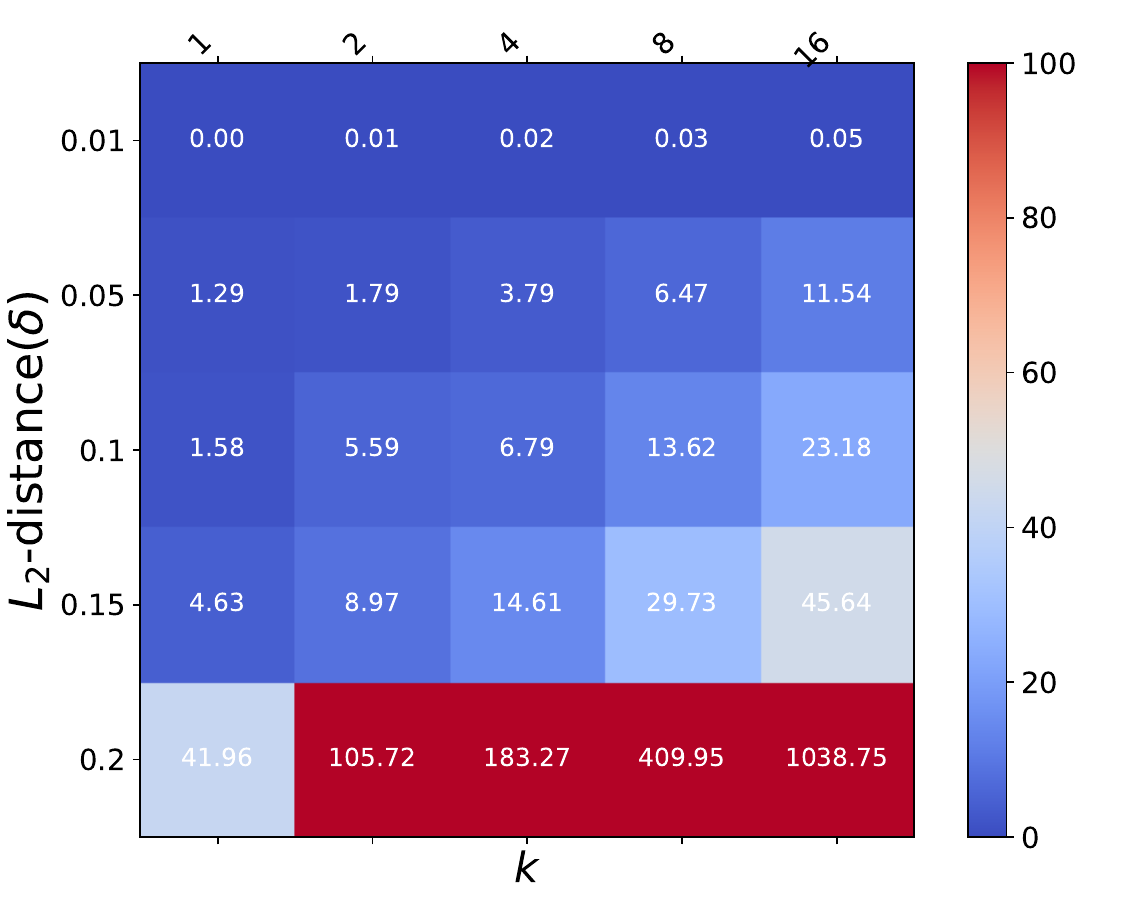}
    \caption{Textual Inversion}
    \label{fig:DE_sub5}
  \end{subfigure}%
  \begin{subfigure}{.3\linewidth}
    \centering
    \includegraphics[width=\linewidth]{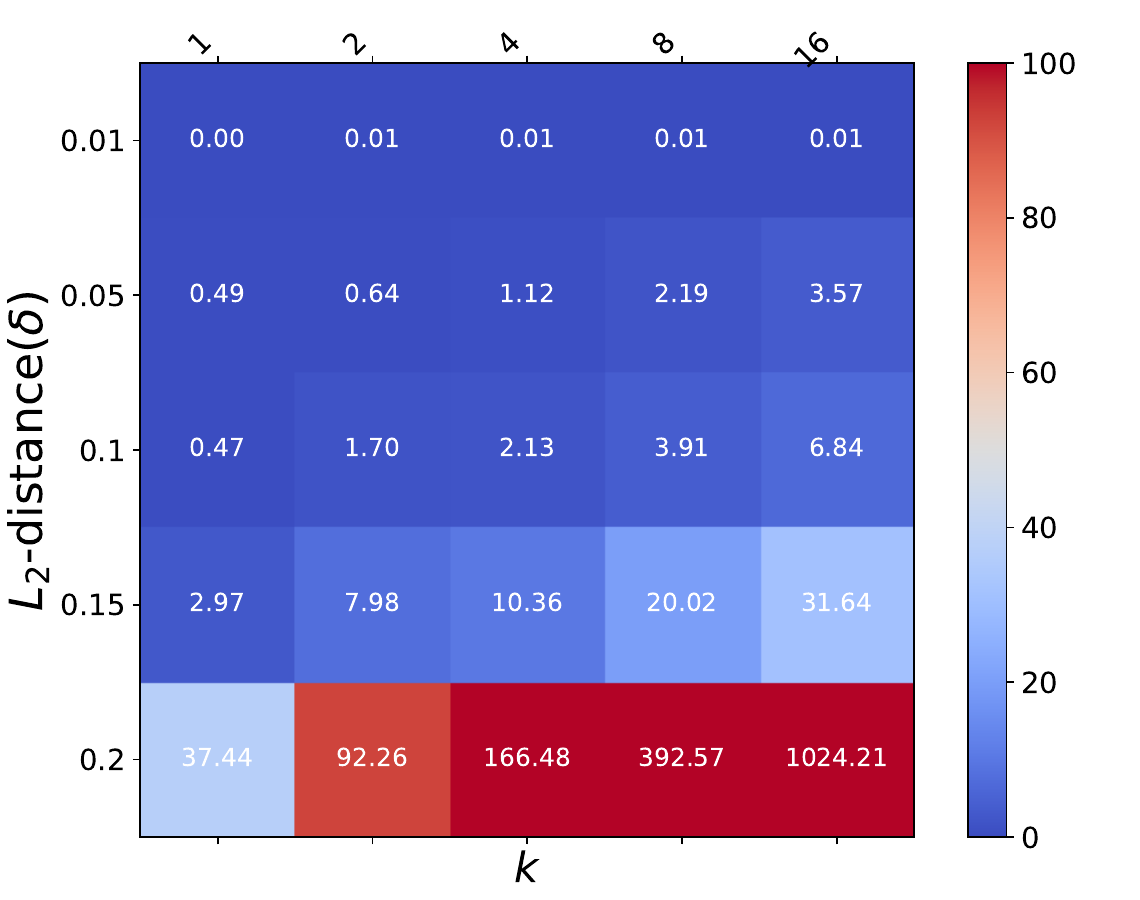}
    \caption{Hypernetwork}
    \label{fig:DE_sub6}
  \end{subfigure}\\
  \begin{subfigure}{.3\linewidth}
    \centering
    \includegraphics[width=\linewidth]{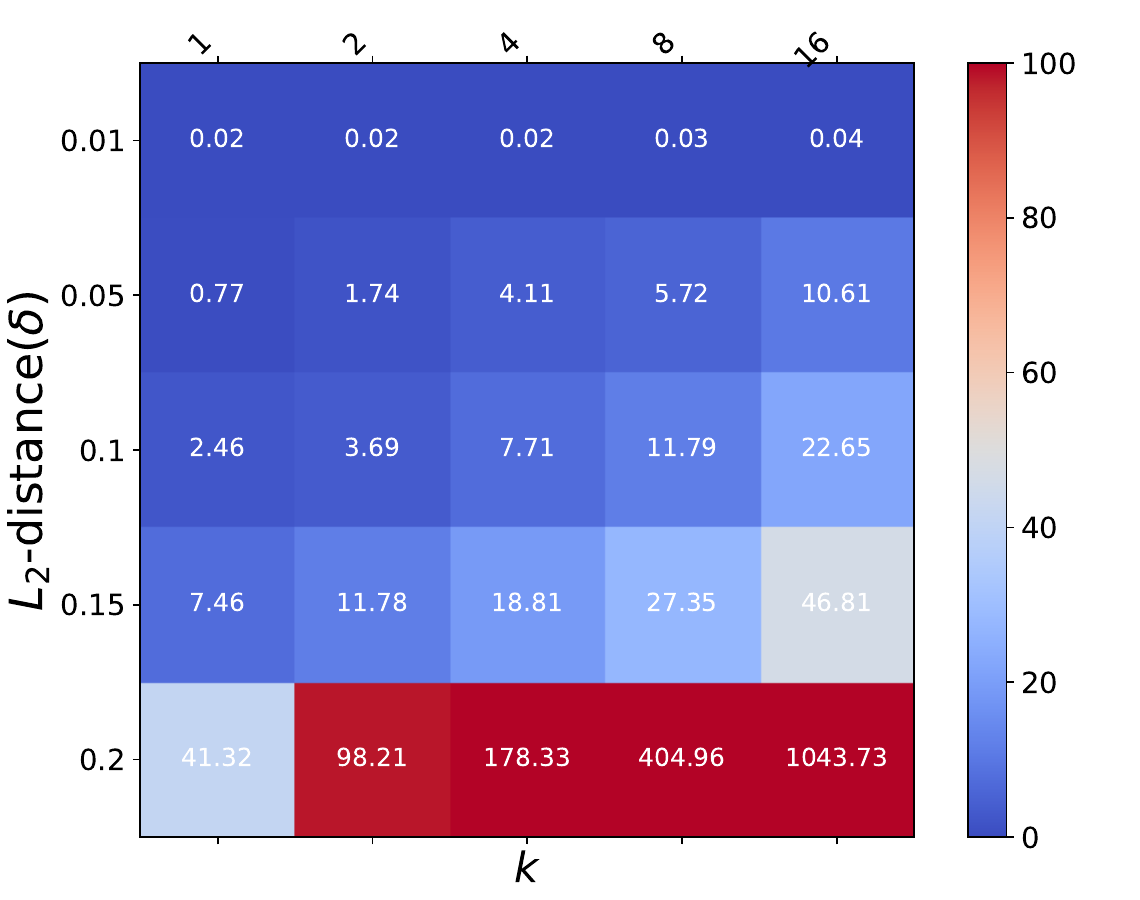}
    \caption{LoRA}
    \label{fig:DE_sub7}
  \end{subfigure}%
  \begin{subfigure}{.3\linewidth}
    \centering
    \includegraphics[width=\linewidth]{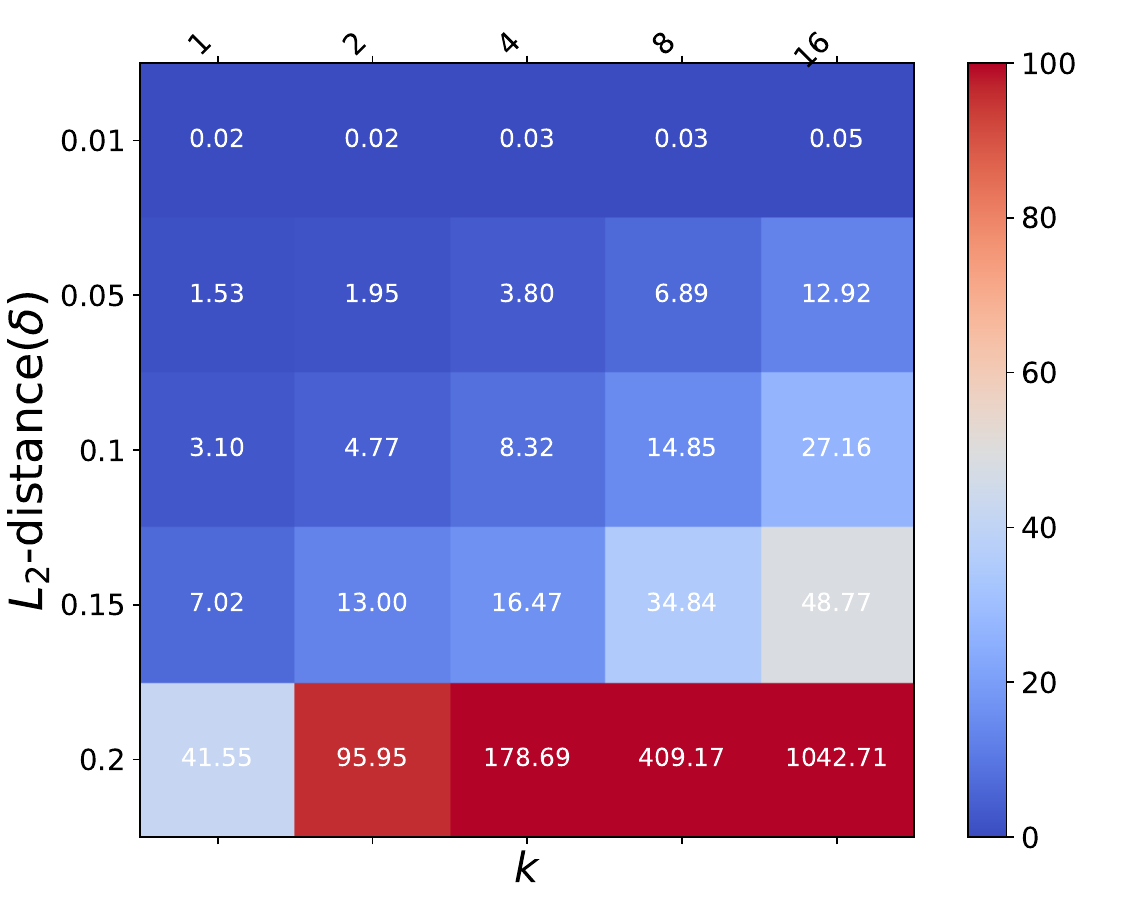}
    \caption{DreamBooth+Hypernetwork}
    \label{fig:DE_sub8}
  \end{subfigure}%
  \begin{subfigure}{.3\linewidth}
    \centering
    \includegraphics[width=\linewidth]{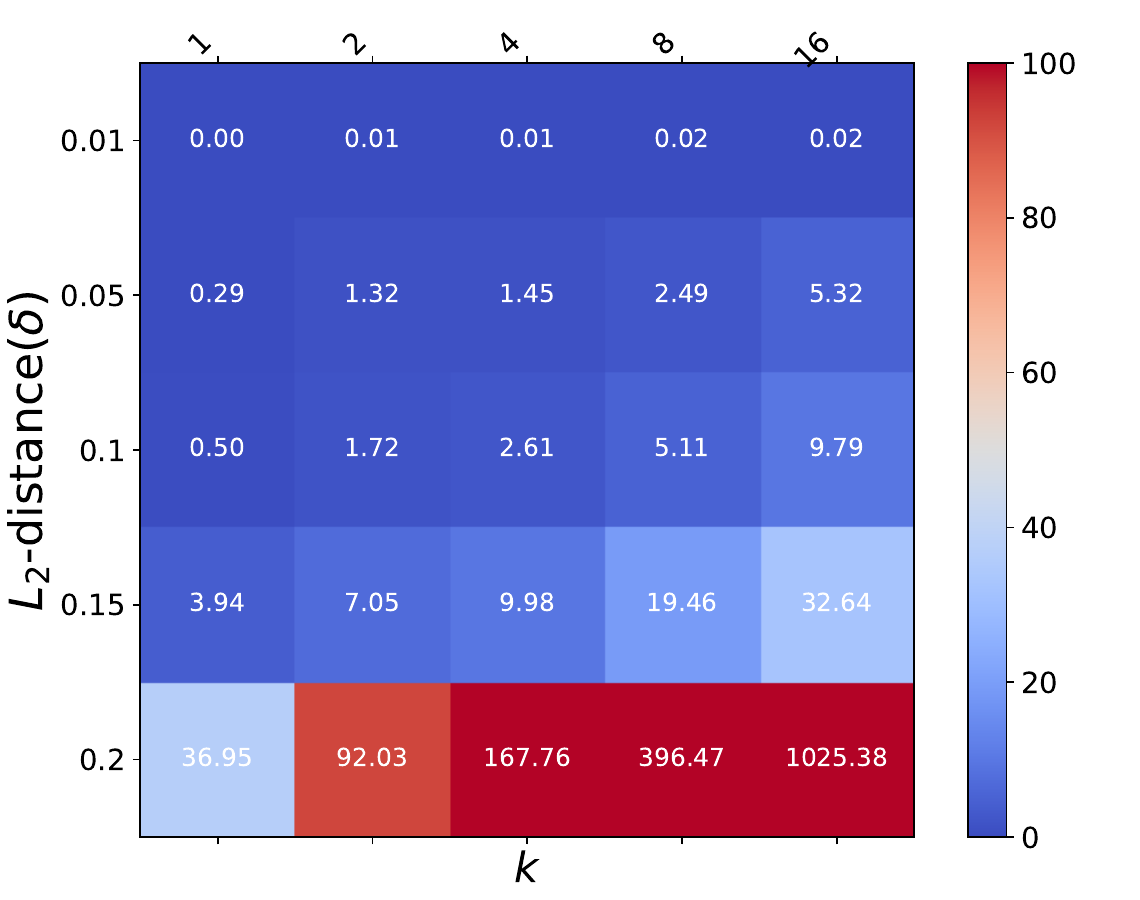}
    \caption{DreamBooth+LoRA}
    \label{fig:DE_sub9}
  \end{subfigure}%
  \caption{The DE results of S2L under variable $L_2$-distance threshold $\delta$ and similar sample number $k$ of the Eidetic memorization. Other experiment settings are kept the same with \cref{tb_main}.}
  \label{fig:DE_grid}
\end{figure*}

In this section, we provide the data extraction results under variable distance threshold $\delta$ and similar sample number $k$ of $(k,l,\delta)$-Eidetic memorization, to better understand how the private samples are memorized. Specifically, \ding{182} by varying the similar sample number $k$ we can see how the data duplication in the pre-training set can affect the data extraction results; \ding{183} by varying the $L_2$-distance threshold $\delta$, we know the data extraction performance at different Eidetic level. The $L_2$-distance threshold $\delta$ is in the range of $[0.01,0.20]$ as we find $\delta>0.20$ to make the extraction algorithms recognize most of the generated images visually irrelevant to their closest images in the pre-training set as successful extractions. The similar sample number $k$ is in the range of $[1,16]$. We keep other experiment settings the same with \cref{tb_main}, and the results are shown in \cref{fig:DE_grid}.

Overall, we find that the extracted example number grows proportionally with the $L_2$-distance $\delta$. Interestingly, we find that after S2L fine-tuning, there is a non-trivial number of extracted samples with few duplications in the pre-training dataset. For example, when $k=1$ and $\delta=0.15$, S2L increases the extracted example number from 0.00 to a range of 0.47 to 3.23; when $k=2$ and $\delta=0.10$, S2L increases the extracted example number from 0.00 to a range of 1.70 to 5.59. This means that in the target domains, S2L can ``shake out" examples that are seen very few times during training. 

\begin{table*}[]
\begin{tabular}{@{}cccc@{}}
\toprule
\textbf{Method} & \textbf{Fine-tuning Set} & \textbf{MIA AUC} & \textbf{Data Extraction} \\ \midrule
Pre-trained     & —                        & 0.712            & 0                        \\
End-2-End       & OOD                      & 0.682            & 0                        \\
S2L             & SP set                   & \textbf{0.766}   & \textbf{15.8}            \\ \bottomrule
\end{tabular}
\end{table*}

\end{document}

%% file: macros.tex
\usepackage{amsthm}
\usepackage{amssymb}
\usepackage{mathtools}
\usepackage{hyperref}
\usepackage[nameinlink,capitalize]{cleveref}
\hypersetup{colorlinks=true,linkcolor=blue,citecolor=blue,urlcolor=blue,pdfborder={0 0 0}}
\usepackage[normalem]{ulem} 
\usepackage{mathtools}

\usepackage[utf8]{inputenc} 
\usepackage[T1]{fontenc}    
\usepackage{hyperref}       
\usepackage{url}            
\usepackage{booktabs}       
\usepackage{amsfonts}       
\usepackage{nicefrac}       
\usepackage{microtype}      
\usepackage{proof-at-the-end}  
\usepackage{multirow}
\usepackage{lscape}         
\usepackage{subcaption}

\usepackage{graphicx,wrapfig}
\usepackage{algorithm}
\usepackage[noend]{algpseudocode}
\usepackage{amsfonts}
\usepackage{url}
\usepackage{enumitem}
\usepackage{amsthm}
\usepackage{thmtools}
\usepackage{thm-restate}

\theoremstyle{definition}
\newtheorem{definition}{Definition}[section]
\theoremstyle{remark}

\usepackage{pifont}
\newcommand{\colorline}[1]{
\hspace{-0.01\linewidth}\colorbox{gray!20}{\makebox[0.99\linewidth][l]{#1}}
 }

\newcommand{\cA}{{\mathcal{A}}}

\newcommand{\cD}{{\mathcal{D}}}
\newcommand{\cE}{{\mathcal{E}}}

\newcommand{\cM}{{\mathcal{M}}}

\newcommand{\cP}{{\mathcal{P}}}
\newcommand{\cQ}{{\mathcal{Q}}}

\newcommand{\bc}{\begin{center}}
\newcommand{\ec}{\end{center}}

\newcommand{\bdm}{\begin{displaymath}}
\newcommand{\edm}{\end{displaymath}}

\newcommand{\beq}{\begin{equation}}
\newcommand{\eeq}{\end{equation}}

\newcommand{\bfl}{\begin{flushleft}}
\newcommand{\efl}{\end{flushleft}}

\newcommand{\bt}{\begin{tabbing}}
\newcommand{\et}{\end{tabbing}}

\newcommand{\beqn}{\begin{align}}
\newcommand{\eeqn}{\end{align}}

\newcommand{\beqs}{\begin{align*}} 
\newcommand{\eeqs}{\end{align*}}  